\documentclass[journal]{IEEEtran}

\PassOptionsToPackage{numbers, compress}{natbib}

\usepackage[utf8]{inputenc} %
\usepackage[T1]{fontenc}    %
\usepackage{url}            %
\usepackage{booktabs}       %
\usepackage{dcolumn}        %
\usepackage{amsfonts}       %
\usepackage{nicefrac}       %
\usepackage{microtype}      %
\usepackage{mathrsfs}
\usepackage{amsmath}
\usepackage{amssymb}
\usepackage{mathtools}
\usepackage{graphics}
\usepackage{float}
\usepackage{epstopdf}
\usepackage{graphicx}
\usepackage{bm}
\usepackage{amsthm}
\usepackage{tikz}
\usepackage[lined,boxed,commentsnumbered,linesnumbered, ruled]{algorithm2e}
\usepackage{comment}
\usepackage{balance}
\usepackage{cite}
\usepackage[capitalize]{cleveref}
\usepackage[size=small]{caption}
\usepackage[size=small,labelformat=simple]{subcaption}
\usepackage{ifthen}
\usepackage{csquotes}
\crefname{equation}{\unskip}{\unskip}
\usetikzlibrary{shapes, patterns,decorations.text, decorations.pathreplacing}
\newtheorem{example}{Example}
\newtheorem{remark}{Remark}

\newtheorem{proposition}{Proposition}

\newcommand*{\Scale}[2][4]{\scalebox{#1}{\ensuremath{#2}}}%

\newcommand{\msH}{\mathsf{H}}
\newcommand{\msI}{\mathsf{I}}
\newcommand{\bPsi}{\boldsymbol{\Psi}}
\newcommand{\bPhi}{\boldsymbol{\Phi}}
\newcommand{\bOmega}{\boldsymbol{\Omega}}

\newcommand{\fxp}[2]{\mathbb Q_{\langle #1, #2\rangle}}

\DeclareMathOperator*{\argmin}{arg\,min}
\DeclareMathOperator*{\supp}{supp}

\definecolor{darkgreen}{rgb}{0, 0.625, 0}

\title{\huge{CodedPaddedFL and CodedSecAgg: Straggler Mitigation and Secure Aggregation in Federated Learning}}

\author{Reent~Schlegel,~\IEEEmembership{Student Member,~IEEE}, Siddhartha~Kumar, 
        Eirik~Rosnes,~\IEEEmembership{Senior Member,~IEEE},
        and~Alexandre~Graell~i~Amat,~\IEEEmembership{Senior Member,~IEEE}%

\thanks{This work was financially supported by the Swedish Research Council under grant 2020-03687. Parts of the material in this paper were presented at the IEEE International Conference on Communications (ICC), Seoul, South Korea, May 2022 \cite{Kumar22}, and at the 30th European Signal Processing Conference (EUSIPCO), Belgrade, Serbia, August/September 2022 \cite{Schlegel22}.}
\thanks{R. Schlegel, S. Kumar, and E. Rosnes are with Simula UiB, Bergen, Norway, e-mail: \{reent,~kumarsi,~eirikrosnes\}@simula.no.}%

\thanks{A. Graell i Amat is with the Department of Electrical Engineering, Chalmers University of Technology, Gothenburg, Sweden, e-mail: alexandre.graell@chalmers.se, and with Simula UiB, Bergen, Norway.}}%

\begin{document}

\maketitle

\begin{abstract}
We present two novel federated learning (FL) schemes that mitigate the effect of straggling devices by introducing redundancy on the devices’ data across the network. Compared to other schemes in the literature, which deal with stragglers or device dropouts by ignoring their contribution, the proposed schemes do not suffer from the client drift problem. The first scheme, CodedPaddedFL, mitigates the effect of stragglers while retaining the privacy level of conventional FL. It combines one-time padding for user data privacy with gradient codes to yield straggler resiliency. The second scheme, CodedSecAgg, provides straggler resiliency and robustness against model inversion attacks and is based on Shamir's secret sharing. We apply CodedPaddedFL and CodedSecAgg to a classification problem. For a scenario with 120 devices, CodedPaddedFL achieves a speed-up factor of 18 for an accuracy of 95\% on the MNIST dataset compared to conventional FL. Furthermore, it yields similar performance in terms of latency compared to a recently proposed scheme by Prakash \emph{et al.} without the shortcoming of additional leakage of private data. CodedSecAgg outperforms the state-of-the-art secure aggregation scheme LightSecAgg by a speed-up factor of 6.6--18.7 for the MNIST dataset for an accuracy of 95\%.
\end{abstract}

\section{Introduction}

Federated learning (FL) \cite{McM17,Kon16,Tian20} is a distributed learning paradigm that %
trains an algorithm across multiple devices without exchanging  the training data directly, thus limiting the privacy leakage and reducing the communication load. More precisely, FL enables multiple devices to collaboratively learn a global model  under the coordination of a central server. At each epoch, the devices train
a local model on their local data and send the
locally-trained models to the central server. The central server
aggregates the local models to update the global model, which is sent to the devices for the next epoch of the training. FL has been used in real-world applications, e.g.,   for medical data \cite{Joc17}, text predictions on mobile devices \cite{Bon19},  or by Apple to personalize Siri.

Training over many heterogeneous devices can  be detrimental to the overall latency due to the effect of so-called stragglers, i.e.,  devices that take exceptionally long to finish their tasks due to random phenomena such as processes running in the background and memory access. Dropouts, which can
be seen as an extreme case of straggling, may also occur. One of the most common ways to address the straggling/dropout problem in FL is to ignore the result of the slowest devices, such as in  federated averaging \cite{McM17}. However, while this approach has only a small impact on the training accuracy when the data is homogeneous across devices, ignoring updates from the slowest devices can lead to the \emph{client drift} problem when the data is not identically distributed across devices\cite{Cha20,Mit21}\textemdash the global model will  tend toward local solutions of the fastest devices, which impairs the overall accuracy of the scheme. In \cite{Mit21,Xie19,Li19,Li20,Wan20,Wu21}, asynchronous schemes have been proposed for straggler mitigation with non-identically distributed data in which the central server utilizes stale gradients from straggling devices. However, these schemes do not in general converge to the global optimum\cite{Mit21}.

FL is also prone to  model inversion attacks~\cite{FredriksonJhaRistenpart15_1,wang2019},  which allow the central server to  infer information about the local datasets through the local gradients collected in each epoch. To prevent such attacks and preserve users' data privacy, secure aggregation protocols have been proposed \cite{bonawitz2017,Kad20,So20,Bell20,elkordy2022,yang2021,Zhao21,Guowen20,Tayyebeh22,Chowdhury21,Tayyebeh22_1,so2021,Nguyen22}
in which the central server only obtains the sum of all the local model updates instead of the local updates directly.  The schemes in \cite{bonawitz2017,Kad20,So20,Bell20,elkordy2022,yang2021,Zhao21,Guowen20,Tayyebeh22,Chowdhury21,Tayyebeh22_1,so2021,Nguyen22}
provide security against inversion attacks by hiding devices’ local models via masking. The masks have an additive
structure so that they can be removed when aggregated at the
central server. To provide resiliency against stragglers/dropouts, secret sharing of the random seeds that generate the masks between
the devices is performed, so that the central server can cancel the masks belonging to dropped devices. Among these schemes, LightSecAgg \cite{yang2021} is one of the most efficient. The schemes \cite{bonawitz2017,Kad20,So20,Bell20,elkordy2022,yang2021,Zhao21,Guowen20,Tayyebeh22,Chowdhury21,Tayyebeh22_1} ignore the contribution of straggling and dropped
devices. However, ignoring  straggling (or dropped) devices
makes these schemes sensitive to the client drift problem.
The schemes in \cite{so2021,Nguyen22} are asynchronous straggler-resilient schemes that do not in general converge to the global optimum. %

The straggler problem has  been addressed in the neighboring area of distributed computing using tools from coding theory. The key idea is to introduce redundancy on the data via an erasure correcting code before distributing it to the servers so that the computations of a subset of the servers are sufficient to complete the global computation, i.e., the computations of straggling servers can be ignored without loss of information. Coded distributed computing has been proposed for matrix-vector and matrix-matrix multiplication \cite{Li16,Yu2017,Lee18,Sev19,amir2019,Dutta2019,Dutta2020}, distributed gradient descent \cite{Tan17}, and distributed optimization \cite{Kar17}. 

Coding for straggler mitigation has  also been proposed for edge computing \cite{Sch22,Zha19,Fri21} and FL\cite{Pra21}. %
The scheme in \cite{Pra21} lets each device generate parity data on its local data and share it  with the central server. This allows the central server to recover part of the information corresponding to the local gradients of the straggling devices without waiting for their result in every epoch. However, sharing parity data with the central server leaks information and hence the scheme provides a lower privacy level than conventional FL.

In this paper, borrowing tools from coded distributed computing and edge computing, we propose two novel FL schemes, referred to as CodedPaddedFL and CodedSecAgg, that provide resiliency against straggling devices (and hence dropouts) by introducing redundancy on the devices’ local data. Both schemes can be divided into two phases. In the first phase, the devices share an encoded version of their data with other devices. In the second phase, the devices and the central server iteratively and collaboratively train a global model. The proposed schemes achieve significantly lower training latency than state-of-the-art schemes.

Our main contributions are summarized as follows.
\begin{itemize}
    \item We present CodedPaddedFL, an FL scheme that provides resiliency against straggling devices  while retaining the same level of privacy as conventional FL. CodedPaddedFL  combines one-time padding to yield privacy with gradient codes \cite{Tan17} to provide straggler resilience.
    Compared to the recent scheme in~\cite{Pra21}, which also exploits erasure correcting codes to yield straggler resiliency, the proposed scheme does not leak additional information.
    
    \item We present CodedSecAgg, a secure aggregation scheme that provides straggler resiliency
    by introducing redundancy on the devices’ local data via Shamir’s secret sharing. CodedSecAgg provides information-theoretic security against model inversion up to a given number of colluding malicious agents (including the central server).   CodedPaddedFL and CodedSecAgg  provide convergence to the true global optimum and hence do not suffer from the client drift phenomenon.

    \item For both schemes, we introduce a  strategy for grouping the devices that significantly reduces the initial latency due to the sharing of the data as well as the decoding complexity at the central server at the expense of a slightly reduced straggler mitigation capability.

        \item Neither one-time padding nor secret sharing can be applied to real-valued data. To circumvent this problem, the proposed schemes are based on a fixed-point arithmetic representation of the real data and subsequently fixed-point arithmetic operations.

\end{itemize}

To the best of our knowledge, our work is the first to apply coding ideas to mitigate the effect of stragglers in FL  without leaking additional information.

The proposed schemes are tailored to linear regression. However, they can be
applied to nonlinear models via kernel embedding. We apply CodedPaddedFL and CodedSecAgg to a classification problem on the MNIST\cite{Cun10} and Fashion-MNIST\cite{Xia17} datasets. 
For a scenario with $120$ devices, CodedPaddedFL achieves a speed-up factor of $18$  on the MNIST dataset for an accuracy of $95$\%  compared to conventional FL, while it shows similar performance in terms of latency compared to the scheme in \cite{Pra21} without leaking additional data.  %
CodedSecAgg achieves a speed-up factor of $6.6$ for $60$ colluding agents and up to $18.7$
for a single malicious agent compared to LightSecAgg for an
accuracy of $95$\%  on the MNIST dataset. 
Our numerical results include  the impact of the decoding in the overall latency, which is often neglected in the literature (thus making comparisons unfair as the decoding complexity may have a significant impact on the global latency \cite{Sev19}).

\section{Preliminaries}

\subsection{Notation}

 We use uppercase and lowercase bold letters for matrices and vectors, respectively, italics for sets, and sans-serif letters for random variables, e.g., \(\bm X\), \(\bm x\),  \(\mathcal{X}\), and \(\mathsf{X}\)  represent a matrix, a vector, a set, and a random variable, respectively. An exception to this rule is $\bm \epsilon$, which will denote a matrix. Vectors are represented as row vectors throughout the paper. For natural numbers  \(c\) and \(d\), \(\bm 1_{c\times d}\) denotes an all-one matrix of size \(c\times d\). The transpose of a matrix \(\bm X\) is denoted as \(\bm X^\top\). The support of a vector $\bm x$ is denoted by $\supp(\bm x)$, while the gradient of a function $f(\bm X)$ with respect to $\bm X$ is denoted by $\nabla_{\bm X}f(\bm X)$. Furthermore, we represent the Euclidean norm of a vector $\bm x$ by $\Vert \bm x \Vert$, while the Frobenius norm of a matrix \(\bm X\) is denoted by   \(\Vert\bm X\Vert_\text{F}\).  Given integers \(a, b\in\mathbb{Z}\), \(a<b\), we define  \([a,b]\triangleq\{a,\ldots,b\}\), where \(\mathbb{Z}\) is the set of integers, and \([a]\triangleq\{1,\ldots, a\}\) for a positive integer $a$. Additionally, we use  \((a)_b\) as a shorthand notation for \(a \bmod{b}\).
 For a real number $e$, $\lfloor e \rfloor$ is the largest integer less than or equal to $e$ and $\lceil e\rceil$ is the smallest integer larger than or equal to $e$.  The expectation of a random variable $\mathsf{\Lambda}$ is denoted by $\mathbb{E}[{\mathsf\Lambda}]$, and we write $\mathsf{\Lambda} \sim  \text{geo}(1-p)$ to denote that $\mathsf{\Lambda}$ follows a geometric distribution with failure probability \(p\). $\msI(\cdot;\cdot)$ denotes the mutual information and $\msH(\cdot|\cdot)$ the conditional entropy.

\subsection{Fixed-Point Numbers} \label{sec:Fixed-Point Numbers}
Fixed-point numbers are rational numbers with a fixed-length integer part and a fixed-length fractional part. A fixed-point number with length $k$ bits and resolution $f$ bits can be seen as an integer from $\mathbb Z_{\langle k \rangle}=[-2^{k-1}, 2^{k-1}-1]$ scaled by $2^{-f}$. In particular, for fixed-point number $\tilde{x}$ it holds that $\tilde{x} = \bar{x}\cdot 2^{-f}$ for some $\bar{x}\in\mathbb Z_{\langle k \rangle}$. We define the set of all fixed-point numbers with length $k$ and resolution $f$ as $\fxp{k}{f}\triangleq\{\tilde x =\bar x 2^{-f}, \bar x \in \mathbb Z_{\langle k \rangle}\}$. The set $\fxp{k}{f}$ is used to represent real numbers in the interval between $-2^{k-f-1}$ and $2^{k-f-1}$ with a finite amount of, i.e. $k$, bits.

\subsection{Cyclic Gradient Codes}
\label{sec:gradcodes}
Gradient codes\cite{Tan17} are a class of codes that have been suggested for straggler mitigation in distributed learning and work as follows. A central server encodes partitions of training data via a gradient code. These coded partitions are assigned to servers which perform gradient computations on the assigned coded data. The central server is then able to decode the sum of the gradients of all  partitions by contacting only a subset of the servers. In particular, a gradient code that can tolerate $\beta - 1$ stragglers in a scenario with $\gamma$ servers and $\gamma$ partitions encodes $\gamma$ partitions into $\gamma$ codewords, one for each server, such that a linear combination of any $\gamma-\beta+1$ codewords yields the sum of all gradients of all partitions. We will refer to such a code as a $(\beta,\gamma)$ gradient code.
A $(\beta, \gamma)$ gradient code over $\fxp{k}{f}$ consists of an encoding matrix $\bm B\in\fxp{k}{f}^{\gamma\times \gamma}$ and a decoding matrix $\bm A\in\fxp{k}{f}^{S\times \gamma}$, where $S$ is the number of straggling patterns the central server can decode.  The encoding matrix $\bm B$ has a cyclic structure and the support of  each row is of size $\beta$, while the the support of each row of the decoding matrix $\bm A$ is of size $\gamma-\beta +1$. The support of the $i$-th row of $\bm B$ dictates which partitions are included in the codeword at server $i$, and the entries of the $i$-row  are the coefficients of the linear combination of the corresponding partitions at server $i$. Let $\bm g_1,\ldots, \bm g_\gamma$ be the gradients on partition $1,\dots,\gamma$. Then, the gradient computed by server $i$ is given by the $i$-th row of $\bm B\left(\bm g_1^\top,\ldots, \bm g_\gamma^\top\right)^\top$.
The central server waits for the gradients of the $\gamma-\beta+1$ fastest servers to decode. Let $\mathcal A$ be the index set of these fastest devices. The central server picks the row of $\bm A$ with support $\mathcal A$ and applies the linear combination given by this row on the received gradients. In order for the central server to receive $\sum_i \bm g_i$, the requirements on $\bm A$ and $\bm B$ are 
\begin{align}
    \label{Eq: GradientCodeCondition}
    \bm A\bm B=\bm 1_{S\times \gamma}\,.
\end{align}
The construction of $\bm A$ and $\bm B$ can be found in \cite[Alg.~1]{Tan17} and \cite[Alg.~2]{Tan17}, respectively.

\subsection{Shamir's Secret Sharing Scheme}
\label{sec:shamir}
Shamir's secret sharing scheme (SSS)\cite{Shamir} over some field $\mathbb{F}$ with parameters $(n',k')$ encodes a secret $x\in\mathbb F$  into $n'$ shares $s_1,\dots,s_{n'}$ such that the mutual information between $x$ and any set of less than $k'$ shares is zero, while any set of $k'$ or more shares contain sufficient information to reconstruct the secret $x$. More precisely, for any $\mathcal{I} \subset \{s_1,\dots,s_{n'}\}$ with $|\mathcal I| < k'$ and any $\mathcal{J} \subseteq \{s_1,\dots,s_{n'}\}$ with $|\mathcal{J}| \geq k'$, we have $\msI(x;\mathcal{I}) = 0$ and $\msH(x|\mathcal{J})=0$.

Shamir's SSS achieves these two properties by encoding $x$ together with $k'-1$ independent and uniformly random samples $\mathsf{r}_1,\dots,\mathsf{r}_{k'-1}$ using a nonsystematic $(n',k')$ Reed-Solomon code. As a result, any subset of Reed-Solomon encoded symbols, i.e., shares, of size less than $k'$ is independently and uniformly distributed. This means that these shares do not reveal any information about $x$, i.e., $\msI(x;\mathcal{I}) = 0$. On the other hand, the maximum distance separable  property of Reed-Solomon codes guarantees that any $k'$ coded symbols are sufficient to recover the initial message, i.e., $\msH(x|\mathcal{J})=0$, where $\mathcal{J}$ denotes the set of the $k'$ coded symbols.

\section{System Model}

In this paper, we consider a network of $D$ devices and a central server. Each device $i$ owns local data $\mathcal{D}_i=\bigl\{(\bm x_j^{(i)}, \bm y_j^{(i)})\mid  j\in[n_i]\bigr\}$ consisting of $n_i$ points with feature vectors $\bm x_j^{(i)}$ and labels $\bm y_j^{(i)}$. The devices wish to collaboratively train a global linear model $\bm \Theta$ with the help of the central server on everyone's data, consisting of $m=\sum_i n_i$ points in total. The model $\bm \Theta$ can be used to predict a label $\bm y$ corresponding to a given feature vector $\bm x$ as $\bm y=\bm x\bm \Theta$.
Our proposed schemes rely on one-time padding and secret sharing, both of which can not be applied on real-valued data. To circumvent this shortcoming we use a fixed-point representation of the data. In particular, we assume $\bm x_j^{(i)}\in\fxp{k}{f}^d$ and $\bm y_j^{(i)}\in\fxp{k}{f}^c$, where $d$ is the size of the feature space and $c$ the dimension of the label. Note that practical systems often operate in fixed-point representation, hence our schemes do not incur in a limiting assumption.

We represent the data in matrix form  as
\begin{align*}
    \bm X^{(i)} = \left(\begin{matrix}
        \bm x_1^{(i)}\\
        \vdots \\
        \bm x_{n_i}^{(i)}
    \end{matrix}\right)\;\,\text{and}\;\,
    \bm Y^{(i)}= \left(\begin{matrix}
        \bm y_1^{(i)}\\
        \vdots\\
        \bm y_{n_i}^{(i)}
    \end{matrix}
    \right)\,.
\end{align*}
The devices try to infer the global model $\bm{\Theta}$ using federated gradient descent, which we describe next.

\subsection{Federated Gradient Descent}
\label{sec:FedGradDes}

For convenience, we collect the whole data (consisting of \(m\) data points) in matrices $\bm X$ and $\bm Y$  as
\begin{align*}
    \bm X=\left(\begin{matrix}
        \bm x_1\\
        \vdots\\
        \bm x_m
    \end{matrix}\right)=\left(\begin{matrix}
        \bm X^{(1)}\\
        \vdots\\
        \bm X^{(D)}
    \end{matrix}\right)
    \;\,\text{and}\;\,
    \bm Y=\left(\begin{matrix}
        \bm y_1\\
        \vdots\\
        \bm y_{m}
    \end{matrix}\right)
    =\left(\begin{matrix}
        \bm Y^{(1)}\\
        \vdots\\
        \bm Y^{(D)}
    \end{matrix}\right)\,,
\end{align*}
where $\bm X$ is of size $m\times d$ and $\bm Y$  of size  $m\times c$. 
The global model $\bm \Theta$ can be found as the solution of the following minimization problem:
\begin{equation*}
    \bm \Theta = \argmin_{\bm \Theta'}\; f(\bm \Theta')\,,
\end{equation*}
where $f(\bm \Theta)$ is the \emph{global} loss function 
\begin{equation}
    \label{eq:ThetaDefinition}
    f(\bm \Theta) \triangleq\frac{1}{2m}\sum_{l=1}^m \left\Vert \bm x_l\bm \Theta-\bm y_l \right\Vert^2+\frac{\lambda}{2}\left\Vert\bm \Theta\right\Vert^2_\text{F}\,,
\end{equation}
where $\lambda$ is the regularization parameter.

Let 
\begin{align*}
    f_i(\bm \Theta)=\frac{1}{2n_i}\sum_{j=1}^{n_i}\Vert\bm x_j^{(i)}\bm \Theta- \bm y_j^{(i)}\Vert^2
\end{align*}
be the \emph{local} loss function at device $i$. We can then write the global loss function in \cref{eq:ThetaDefinition} as 
\begin{align*}
f(\bm \Theta)=\sum_{i=1}^{D}\frac{n_i}{m}f_i(\bm \Theta) + \frac{\lambda}{2}\left\Vert\bm \Theta\right\Vert^2_\text{F}\,.
\end{align*}
In federated gradient descent, the model $\bm{\Theta}$ is trained iteratively over multiple epochs on the local data at each device. At each epoch, the devices compute the gradient on the local loss function of the current model and send it to the central server. The central server then aggregates the local gradients to obtain a global gradient which is used to update the model.  
More precisely, during the \(e\)-th epoch,  device $i$ computes the gradient 
\begin{align}
    \label{Eq: DeviceComputation}
    \Scale[1.0]{\bm G_i^{(e)}= n_i\nabla_{\bm \Theta} f_i(\bm \Theta^{(e)})=
    {\bm X^{(i)\top}}\bm X^{(i)}\bm \Theta^{(e)}-{\bm X^{(i)\top}} \bm Y^{(i)}}\,,
\end{align}
where $\bm \Theta^{(e)}$ denotes the current model estimate. 
Upon reception of the gradients, the central server aggregates them as $\bm G^{(e)} = \sum_i \bm G_i^{(e)}$ to update the model according to
\begin{align}
    \label{Eq: ServerAggregation}
    \nabla_{\bm \Theta} f(\bm \Theta^{(e)})&=\frac{1}{m}\bm G^{(e)}+\lambda\bm \Theta^{(e)}\,,\\
    \label{Eq: ServerUpdate}
    \bm \Theta^{(e+1)}&=\bm \Theta^{(e)}-\mu\nabla_{\bm \Theta} f(\bm \Theta^{(e)})\,,
\end{align}
where \(\mu\) is the learning rate. The updated model  \(\bm \Theta^{(e+1)}\) is then sent back to the devices, and \cref{Eq: DeviceComputation,Eq: ServerAggregation,Eq: ServerUpdate} are iterated $E$ times until convergence, i.e., until \(\bm \Theta^{(E+1)}\approx\bm \Theta^{(E)}\).

\subsection{Computation and Communication Latency}
\label{sec:comp_comm_lat}
We model the computation times of the devices as random variables with a shifted exponential distribution, as is common in the literature \cite{Zha19}. This means that the computation times  comprise a deterministic time corresponding to the time a device takes to finish a computation in its processing unit and a random setup time due to unforseen delays such as memory access and other tasks running in the background. Let $\mathsf{T}^{\mathsf{comp}}_i$ be the time it takes device $i$ to perform $\rho_i$ multiply and accumulate (MAC) operations. We then have
\begin{equation*}
    \mathsf{T}^{\mathsf{comp}}_i= \frac{\rho_i}{\tau_i}+\mathsf{\Lambda}_i\,,
\end{equation*}
with $\tau_i$ being the deterministic number of MAC operations device $i$ performs per second and $\mathsf{\Lambda}_i$   the random exponentially distributed setup time with $\mathbb{E}[\mathsf{\Lambda}_i]=1/\eta_i$.

The devices communicate with the central server through a secured, i.e., authenticated and encrypted, wireless link. This communication link is unreliable and may fail. In case a packet is lost, the sender retransmits until a successful transmission occurs. Let $\mathsf{N}_i^{\mathsf{u}}\sim\mathrm{geo}\,(1-p_i)$ and $\mathsf{N}_i^{\mathsf{d}}\sim\mathrm{geo}\,(1-p_i)$ be the number of tries until device $i$ successfully uploads and downloads a packet to or from the central server, and let $\gamma^{\mathsf{u}}$ and $\gamma^{\mathsf{d}}$ be the transmission rates in the upload and download. Then, the time it takes device $i$ to upload or download $b$ bits is
\begin{equation*}
    \mathsf{T}^{\mathsf{u}}_i=\frac{\mathsf{N}^{\mathsf{u}}_i}{\gamma^{\mathsf{u}}}b \quad \text{and} \quad \mathsf{T}^{\mathsf{d}}_i=\frac{\mathsf{N}^{\mathsf{d}}_i}{\gamma^{\mathsf{d}}}b\,,
\end{equation*}
respectively.
Furthermore, we assume that the devices feature full-duplex transmission capabilities, as per the LTE Cat 1 standard for Internet of Things (IoT) devices, and have access to orthogonal channels to the central server. This means that devices can simultaneously transmit and receive messages to and from the central server without interference from the other devices.\footnote{The proposed schemes apply directly to the half-duplex case as well. However, the full-duplex assumption allows to reduce the latency and simplifies its  analysis.}

Last but not least, all device-to-device (D2D) communication is authenticated and encrypted and is routed through the central server. This enables efficient D2D communication, because the routing through the central server guarantees that any two devices in the network can communicate with each other even when they are spatially separated, and the authentication and encryption prohibit man-in-the-middle attacks and eavesdropping.

\subsection{Threat Model and Goal}
We assume a scenario where  the central server and the  devices are honest-but-curious. 
The goal of CodedPaddedFL is to provide straggler resiliency while achieving  the same level of privacy as conventional FL, i.e., the central server does not gain additional information compared to conventional FL,  and  (colluding) devices do not gain any information on the data shared by other devices.  For the secure aggregation scheme, CodedSecAgg, we assume that up to $z$ agents (including the central server) may  collude to infer information about the
local datasets of other devices. The goal is to ensure  device data privacy against the $z$ colluding agents while providing
straggler mitigation. Privacy
in this setting means that malicious devices do not gain any information about the local datasets of other devices and that the central server only learns the aggregate of all local gradients to prevent a model inversion attack.

\section{Privacy-Preserving Operations on Fixed-Point Numbers}
\label{sec:priv_op_fp}

CodedPaddedFL, introduced in the next section, is based  on one-time padding to provide data privacy, while CodedSecAgg, introduced in Section~\ref{eq:CodedSecAgg}, is based on Shamir's SSS. As mentioned before, neither  one-time padding nor secret sharing can be applied over real-valued data. Hence, we resort to using a fixed-point representation of the data. In this section, we explain how to perform elementary operations on fixed-point numbers.

Using fixed-point representation of data for privacy-preserving computations was first introduced in \cite{Cat10} in the context of multi-party computation. The idea is to use the integer $\bar{x}$ to represent the fixed-point number $\tilde{x} = \bar{x}\cdot 2^{-f}$. To this end, the integer $\bar{x}$ is mapped into a finite field and can then be secretly shared with other devices to perform secure operations such as addition, multiplication, and division with other secretly shared values. In CodedPaddedFL and CodedSecAgg we will use a similar approach. However, %
we only need to multiply a known number with a padded number and not two padded numbers with each other, which significantly simplifies the operations.

Consider the  fixed-point datatype $\fxp{k}{f}$ (see \cref{sec:Fixed-Point Numbers}). Secure addition on $\fxp{k}{f}$ can be performed via simple integer addition with an additional modulo operation. Let $(\cdot)_{\mathbb Z_{\langle k \rangle}}$ be the map from the integers onto the set $\mathbb Z_{\langle k \rangle}$ given by the modulo operation. Furthermore, let $\tilde{a},\tilde{b}\in\fxp{k}{f}$, with $\tilde{a} = \bar{a}2^{-f}$ and $\tilde{b} = \bar{b}2^{-f}$. For $\tilde{c} = \tilde{a} + \tilde{b}$, with $\tilde{c} = \bar{c}2^{-f}$, we have $\bar{c} = (\bar{a} + \bar{b})_{\mathbb Z_{\langle k \rangle}}$.

Multiplication on $\fxp{k}{f}$ is performed via integer multiplication with scaling over the reals in order to retain the precision of the datatype and an additional modulo operation. For $\tilde{d} = \tilde{a} \cdot \tilde{b}$, with $\tilde{d} = \bar{d}2^{-f}$, we have $\bar{d} = (\lfloor\bar{a}\cdot\bar{b}\cdot 2^{-f}\rfloor)_{\mathbb Z_{\langle k \rangle}}$.

\begin{proposition}[Perfect privacy]
    \label{Th: Perfect Privacy}
    Consider a secret \(\tilde{x}\in\fxp{k}{f}\) and a one-time pad \(\tilde r\in\fxp{k}{f}\) that is picked uniformly at random. Then, \(\tilde{x}+\tilde{r}\) is uniformly distributed in \(\fxp{k}{f}\), i.e., \(\tilde{x}+\tilde{r}\) does not reveal any information about \(\tilde{x}\).
\end{proposition}

\cref{Th: Perfect Privacy} is an application of a one-time pad, which was proven secure by Shannon in \cite{Shannon49}. It follows that given that an adversary (having unbounded computational power) obtains the sum of the secret and the pad, \(\tilde{x}+\tilde{r}\), and does not know the pad \(\tilde{r}\),  it cannot determine the secret \(\tilde x\).

\begin{proposition}[Retrieval]
    Consider a public fixed-point number \(\tilde{c}\in\fxp{k}{f}\), a secret \(\tilde{x}\in\fxp{k}{f}\), and a one-time pad \(\tilde{r}\in\fxp{k}{f}\) that is picked uniformly at random. Suppose we have the weighted sum \(\tilde{c}(\tilde{x}+\tilde{r})\) and the one-time pad. Then, we can retrieve \(\tilde{c}\tilde{x}=\tilde{c}(\tilde{x}+\tilde{r})-\tilde{c}\tilde{r}+O(2^{-f})\).
\end{proposition}

The above proposition tells us that, given \(\tilde{c}\), \(\tilde{c}(\tilde{x}+\tilde{r})\), and \(\tilde{r}\), it is possible to obtain an approximation of \(\tilde{c}\tilde{x}\). Moreover, if we choose a sufficiently large  \(f\),  then we can retrieve \(\tilde{c}\tilde{x}\) with negligible error.

\section{Coded Federated Learning}
\label{Sec: CodedFederatedBatchGradient}

In this section, we introduce our first proposed scheme, named CodedPaddedFL. To yield straggler mitigation, CodedPaddedFL is based on the use of gradient codes (see \cref{sec:gradcodes}). More precisely,  each device computes the gradient on a linear combination of the data of a subset of the devices. In contrast to distributed computing, however, where a user willing to perform a computation has all the data available, in an FL scenario the data is inherently distributed across devices and hence gradient codes cannot be applied directly. Thus, to enable the use of gradient codes, we first need to share data between devices. To preserve data privacy, in CodedPaddedFL, our scheme one-time pads the data prior to sharing it.

CodedPaddedFL comprises two phases. In the first phase,  devices share data to enable the use of gradient codes. In the second phase, coded gradient descent is applied on the padded data.\footnote{We remark that our proposed scheme deviates slightly from standard federated gradient descent as described in \cref{sec:FedGradDes} by trading off a pre-computation of the data for more efficient computations at each epoch, as explained in Section~\ref{sec:codedgradientdescent}.}
In the following, we describe both phases.

\subsection{Phase 1: Data Sharing}
\label{sec:CodedPaddedFLPhase1}

In the first phase of CodedPaddedFL, the devices share a one-time padded version of their data with other devices. We explain next how the devices pad their data.

Each device $i$ generates a pair of uniformly random one-time pads $\bm{\mathsf{R}}^{\text{G}}_{i}\in \mathbb Z_{\langle k \rangle}^{d\times c}$ and $\bm{\mathsf{R}}^{\text{X}}_{i}\in \mathbb Z_{\langle k \rangle}^{d\times d}$, with $\bm{\mathsf{R}}^{\text{X}}_{i} = {\bm{\mathsf{R}}^{\text{X}}_{i}}^\top$. Then, device $i$ sends these one-time pads to the central server.
Furthermore, using the one-time pads and its data, device $i$ computes
\begin{align}
    \bm \Psi_{i}&= \bm G_i^{(1)}+\bm{\mathsf{R}}^{\text{G}}_{i}\,, %
    \label{Eq: PaddedData2} \\
    \bm\Phi_{i}&=%
    {\bm X^{(i)\top}} \bm X^{(i)} + \bm {\mathsf{R}}^{\text{X}}_i\,,
    \label{eq:phi}
\end{align}
where $\bm G_i^{(1)}$ is the gradient of device $i$ in the first epoch (see~\cref{Eq: DeviceComputation}). Matrices $\bm \Psi_{i}$ and $\bm \Phi_{i}$ are one-time padded versions of the first gradient and the transformed data. As a result, the mutual information between them and the data at device $i$ is zero. The reason for padding the  gradient of the first epoch in \cref{Eq: PaddedData2} and the transformation of the dataset in \cref{eq:phi} will become clear in \cref{sec:codedgradientdescent}.

Devices then  share the padded matrices $\bm\Psi_{i}$ and $\bm \Phi_{i}$ with $\alpha - 1$ other devices  to introduce redundancy in the network and enable coded gradient descent in the second phase. Particularly, as described in Section~\ref{sec:codedgradientdescent}, each device computes the gradient on a linear combination of a subset of $\{\bPsi_1,\ldots,\bPsi_D\}$ and $\{\bPhi_1,\ldots,\bPhi_D\}$, where the linear combination is determined by an $(\alpha,D)$ gradient code. Let  $\bm A$ and $\bm B$  be the decoding matrix and the encoding matrix of the gradient code, respectively. Each row and column of $\bm B$ has exactly $\alpha$ nonzero elements. The support of row $i$ determines the subset of $\{\bPsi_1,\ldots,\bPsi_D\}$ and $\{\bPhi_1,\ldots,\bPhi_D\}$ on which device $i$ will compute the gradient. Correspondingly, the support of column $j$ dictates the subset of devices with which device $j$ has to share its padded data $\bPsi_j$ and $\bPhi_j$. The cyclic structure of $\bm B$ guarantees that each device will utilize its own data, which is why each device shares its data with only $\alpha -1$ other devices while we have $\alpha$ nonzero elements in each column. 

The sharing of data between devices is  specified by an $\alpha\times D$ assignment matrix $\bOmega$ whose $i$-th column  corresponds to the support of the $i$-th row of matrix $\bm B$. Matrix $\bOmega$ is given by
\begin{align*}
    \bm \Omega=\Scale[0.93]{\left(\begin{matrix}
        1 & 2 & \cdots & D\\
        2 & 3 & \cdots & 1\\
        \vdots & \vdots & \ddots & \vdots\\
        (\alpha-1)_D+1 & (\alpha)_D+1 & \cdots & (\alpha-2)_D+1
    \end{matrix}\right)}\,.
\end{align*}
The entry    at row $i$ and column $j$ of $\bm \Omega$, $\omega_{ij}$, identifies a device sharing its padded data with  device $j$, e.g., $\omega_{\cdot,a}=b$ means that device $b$ shares its data with device $a$. 

\begin{example}\label{ex:omega}
    Consider \(D=3\) devices and \(\alpha=2\). We have the transmission matrix $\bm \Omega=\Scale[0.93]{\left(\begin{matrix}
        1 & 2 & 3\\
        2 & 3 & 1\\
    \end{matrix}\right)}$,
    where, for instance, \(\omega_{21}=2\) denotes that  device $2$ shares its padded gradient and data, \(\bm \Psi_{2}\) and \(\bm\Phi_{2}\), with  device $1$. The first row says that each device  should share its data with itself, making communication superfluous, whereas the second row says that devices \(2, 3\), and \(1\) should share their padded gradients and data with devices \(1, 2\), and \(3\), respectively. 
\end{example}

After the sharing of the padded data, the devices locally encode the local data and the received data using the gradient code. Let $\{b_{i,j}\}$ be the entries of the encoding matrix $\bm B$. Device $i$ then computes
\begin{align}
    \bm C_{i} &=\left(
        b_{i,\omega_{1i}}, \ldots, b_{i, \omega_{\alpha i}}
    \right)\Scale[1.0]{\left(
        \bm \Psi_{\omega_{1i}}^\top,
        \ldots,
        \bm \Psi_{\omega_{\alpha i}}^\top
    \right)^\top}\,,
    \label{eq:C}\\
    \bar{\bm C}_{i} &=\left(
        b_{i,\omega_{1i}}, \ldots, b_{i, \omega_{\alpha i}}
    \right)\Scale[0.96]{\left(
        \bm \Phi_{\omega_{1i}}^\top,
        \ldots, 
        \bm \Phi_{\omega_{\alpha i}}^\top
    \right)^\top}\,,
    \label{eq:Cbar}
\end{align}
where \cref{eq:C} corresponds to the encoding, via the gradient code, of the padded gradient of device $i$ at epoch $1$ and  the padded gradients (at epoch $1$) received from the $\alpha-1$ other devices, and \cref{eq:Cbar} corresponds to the encoding of the padded data of device $i$ as well as the padded data received from the other devices. This concludes the sharing phase of CodedPaddedFL.

\subsection{Phase 2: Coded Gradient Descent}
\label{sec:codedgradientdescent}

In the second phase of CodedPaddedFL, the devices and the central
server  collaboratively   and  iteratively  train  the   global  model
$\bm{\Theta}$.  As the  training is  an iterative  process, the  model
changes in each  epoch. Let $\bm{\Theta}^{(e)}$ be the  model at epoch
$e$. We can write $\bm{\Theta}^{(e)}$ as
\begin{equation}
    \bm \Theta^{(e)}=\bm \Theta^{(1)}+\bm\epsilon^{(e)}\,,
    \label{eq:theta}
\end{equation}
where $\bm\epsilon^{(e)}$ is an update matrix and $\bm \Theta^{(1)}$ is the initial model estimate in the first epoch. In contrast to the standard approach in gradient descent, where the central server sends $\bm \Theta^{(e)}$ to the devices in every epoch, we will use the update matrix $\bm\epsilon^{(e)}$ instead.

When device $i$ receives $\bm \epsilon^{(e)}$, it computes the gradient $\tilde{\bm G}_i^{(e)}$ on the encoded data $\bm C_{i}$ and $\bar{\bm C}_{i}$. More precisely, in epoch $e$ device $i$ computes 
\begin{align}
\tilde{\bm G}_i^{(e)}  &= \bm C_{i}+ \bar{\bm C}_{i} \bm \epsilon^{(e)}\label{eq:codedpaddedFLdevicecomputations}\\
&\overset{(a)}{=} \sum_{j=1}^\alpha b_{i, \omega_{ji}}\bigg(\bm G_{\omega_{ji}}^{(1)} + \bm{\mathsf{R}}^{\text{G}}_{\omega_{ji}}\bigg)\nonumber\\
&\quad\quad+\sum_{j=1}^\alpha b_{i, \omega_{ji}} \bigg({\bm X^{(\omega_{ji})\top}} \bm X^{(\omega_{ji})} + \bm {\mathsf{R}}^{\text{X}}_{\omega_{ji}} \bigg)\bm \epsilon^{(e)}\nonumber\\
&\overset{(b)}{=} \sum_{j=1}^\alpha b_{i, \omega_{ji}}\bigg(\bm G_{\omega_{ji}}^{(1)} + {\bm X^{(\omega_{ji})\top}} \bm X^{(\omega_{ji})}\bm \epsilon^{(e)}\bigg)\nonumber\\
&\quad\quad+ \sum_{j=1}^\alpha b_{i, \omega_{ji}} \bigg( \bm{\mathsf{R}}^{\text{G}}_{\omega_{ji}} + \bm {\mathsf{R}}^{\text{X}}_{\omega_{ji}}\bm \epsilon^{(e)}\bigg)\nonumber\\
&\overset{(c)}{=} \sum_{j=1}^\alpha b_{i, \omega_{ji}} \bigg(\bm G_{\omega_{ji}}^{(e)}+\bm {\mathsf{R}}^{\text{X}}_{\omega_{ji}}\bm\epsilon^{(e)}+\bm{\mathsf{R}}^{\text{G}}_{\omega_{ji}}\bigg)\,,\nonumber
\end{align}
where $(a)$ follows from \cref{eq:C} and \cref{eq:Cbar} together with \cref{Eq: PaddedData2} and \cref{eq:phi}, $(b)$ is a reordering, and $(c)$ follows from \cref{Eq: DeviceComputation} and \cref{eq:theta}. Subsequently, device $i$ sends the gradient $\tilde{\bm G}_i^{(e)}$ to the central server. The central server waits for the gradients from the $D-\alpha+1$ fastest devices before it starts the decoding process, i.e., the central server ignores the results from the $\alpha -1$ slowest devices, which guarantees resiliency against up to $\alpha - 1$ stragglers. The decoding is based on the decoding matrix $\bm A$. Let $\mathcal{A}\subseteq[D]$, with $|\mathcal{A}|=D-\alpha+1$, be the set of the $D-\alpha+1$ fastest devices. The central server, knowing all one-time pads and $\bm\epsilon^{(e)}$, removes the pads from $\tilde{\bm G}_i^{(e)}$, $\forall i\in\mathcal{A}$, and obtains $\bm P_i^{(e)}\triangleq\sum_{j=1}^\alpha b_{i, \omega_{ji}} \bm G_{\omega_{ji}}^{(e)}$. The next step is standard gradient code decoding. Let $\bm a_s=(a_{s,1}, \ldots, a_{s,D})$ be row $s$ from $\bm A$ such that $\supp(\bm a_s)=\mathcal{A}$, i.e., row $s$ is used to decode the straggling pattern $[D] \backslash \mathcal{A}$. Then,
\begin{equation}
    \label{Eq: CodedPaddedFL_UpdateRule}
    \sum_{i\in\mathcal{A}}a_{s,i}\bm P_i^{(e)}\overset{(a)}{=}\bm G^{(e)}
    \overset{(b)}{=}m\big(\nabla_{\bm \Theta}f(\bm \Theta^{(e)}) - \lambda\bm\Theta^{(e)}\big)\,,
\end{equation}
where $(a)$ follows from the property of gradient codes in \cref{Eq: GradientCodeCondition} and $(b)$ follows from \cref{Eq: ServerAggregation}. Lastly, \(\bm \Theta^{(e+1)}\) is obtained according to \cref{Eq: ServerUpdate} for the next epoch. %
\begin{figure}
    \centering
    \includegraphics[width=\columnwidth]{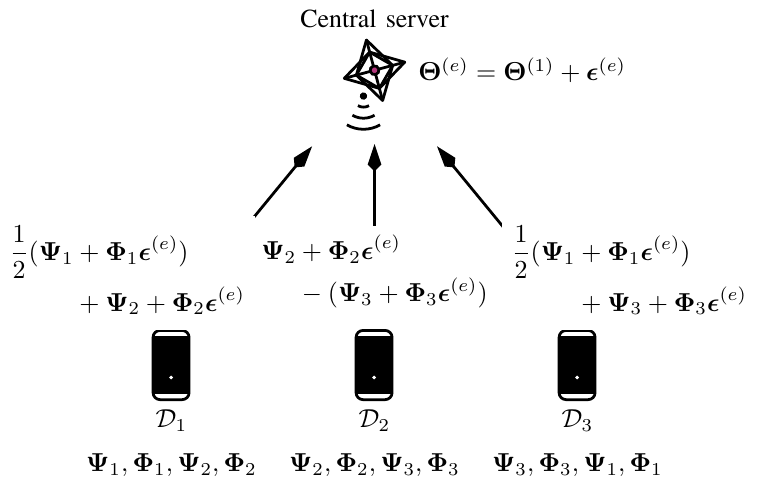}
    \caption{An example showcasing the system model as well as an epoch of the proposed CodedPaddedFL. The system consists of \(D=3\) devices and a central server. The devices share \(\bm \Psi_{i}\) and \(\bm \Phi_{i}\). During the $e$-th epoch, the central server sends \(\bm \epsilon^{(e)}\) to the devices. The devices compute coded gradients using an \((\alpha=2, D)\) gradient code, and send them to the central server, which decodes them to compute the model update.}
    \label{Fig:CFLscheme}
\end{figure}

The proposed CodedPaddedFL  is schematized in Fig.~\ref{Fig:CFLscheme}. It is easy to see that our scheme achieves the global optimum.

\begin{proposition}
\label{prop:Prop}
    The proposed CodedPaddedFL with parameters \((\alpha,D)\)  is resilient to \(\alpha - 1\) stragglers, and achieves the global optimum, i.e., the optimal model obtained through gradient descent for  linear regression.
\end{proposition}
\begin{IEEEproof}
    From \cref{Eq: CodedPaddedFL_UpdateRule}, we see that during each epoch, \(e\), the central server obtains
    \begin{align*}
    \nabla_{\bm \Theta}f(\bm\Theta^{(e)}) &= \frac{1}{m}\bm G^{(e)} + \lambda\bm \Theta^{(e)} \\
    &=\frac{1}{m}\bm X^\top(\bm X\bm\Theta^{(e)} - \bm Y) + \lambda\bm \Theta^{(e)}
    \end{align*}
    using the coded data obtained from the \(D-\alpha+1\) fastest devices. It further obtains an updated linear model using \cref{Eq: ServerUpdate}, which is exactly the update rule for gradient descent.
\end{IEEEproof}

\subsection{Communication Latency of the Data Sharing Phase}
\label{sec:commlatencysharing}
As mentioned in \cref{sec:comp_comm_lat}, we assume that devices are equipped with full-duplex technology and that simultaneous transmission between the $D$ devices and the central server via orthogonal channels is possible. Thus, the sharing of data between devices according to $\bm \Omega$ requires $\alpha - 1$ successive transmissions consisting of upload and download. In particular, the first row of $\bm \Omega$ encompasses no transmission as each device already has access to its own data. For the other rows of $\bm \Omega$, the communication corresponding to the data sharing specified by any given row  can be performed simultaneously, as the devices can communicate in full-duplex with the central server, and each device has a dedicated channel without interference from the other devices. As a result, all data sharing as defined by $\bm \Omega$ is completed after $\alpha - 1$ successive uploads and downloads. Considering \cref{ex:omega}, we can see that $\alpha-1=1$ transmission is enough.
\begin{remark}
    Note that because both $\bm X^{(i)\top}\bm X^{(i)}$ and $\bm{\mathsf{R}}^{\text{X}}_{i}$ are symmetric, $\bm \Phi_i$ is symmetric as well. As a result, device $i$ only has to transmit the upper half of $\bm \Phi_i$ to the other devices.
\end{remark}
We assume the communication cost of transmitting the one-time pads $\bm{\mathsf{R}}^{\text{G}}_i$ and $\bm {\mathsf{R}}^{\text{X}}_i$ to the central server to be negligible as in practice the pads will be generated using a pseudorandom number generator such that it is sufficient to send the much smaller seed of the pseudorandom number generator instead of the whole one-time pads.

\subsection{Complexity}\label{sec:paddedFLComplexity}
We analyze the complexity of the two phases of CodedPaddedFL.

In the sharing phase, when $\alpha > 1$, device $i$ has to upload $\bm \Phi_i$ and $\bm \Psi_i$ to the central server and download $\alpha - 1$ different $\bm \Phi_j$ and $\bm \Psi_j$ from other devices as given by the encoding matrix $\bm B$. As a result, the sharing comprises uploading $d\left(\frac{d+1}{2}+c\right)$  and downloading $\left(\alpha - 1\right)d\left(\frac{d+1}{2}+c\right)$ elements from $\fxp{k}{f}$. The encoding encompasses linearly combining $\alpha$ matrices two times (both $\bm \Phi_i$ and $\bm \Psi_i$). Therefore, each device has to perform $\left(\alpha - 1\right)d\left(\frac{d+1}{2}+c\right)$ MAC operations. In the case of $\alpha = 1$, no sharing of data and encoding takes place.

In the learning phase, devices have to compute a matrix multiplication with a subsequent matrix addition. This requires $d^2c$ MAC operations. Subsequently, the devices transmit their model updates, consisting of $dc$ elements from $\fxp{k}{f}$.

\begin{remark}
During the learning phase, the complexity of CodedPaddedFL is equivalent to the complexity of conventional FL. The computations in \cref{eq:codedpaddedFLdevicecomputations} are as complex as in \cref{Eq: DeviceComputation} given a pre-computation of $\bm X^{(i)\top}\bm X^{(i)}$ and $\bm X^{(i)\top}\bm Y^{(i)}$ and the model updates have the same dimensions.
\end{remark}

\subsection{Grouping}
To yield privacy, our proposed scheme entails a relatively high communication cost in the sharing phase and a decoding cost at the central server, which grow with increasing values of $\alpha$. As we show in Section~\ref{sec:numericalResults}, values of $\alpha$ close to the maximum, i.e., $D$,  yield the lowest overall latency, due to the strong straggler mitigation a high $\alpha$ facilitates. To reduce latency further, one should reduce $\alpha$ while  retaining a high level of straggler mitigation. 
To achieve this, we partition the set of all devices into $N$ smaller disjoint groups and locally apply CodedPaddedFL in each group. It is most efficient to distribute the devices among all $N$ groups as equally as possible, as each group will experience the same latency. However, it is not necessary that $N$ divides $D$.

The central server decodes the aggregated gradients from each group of devices first and obtains the global aggregate as the sum of the individual group aggregates. 

\begin{example}
Assume that there are $D=25$ devices and the central server waits for the $10$ fastest devices to finish their computation. This would result in $\alpha = 16$ as in CodedPaddedFL the central server has to wait for the $D-\alpha +1$ fastest devices. By grouping the devices into $N=5$ groups of $D/N=5$ devices each,  $\alpha = 4$ would be sufficient as the central server has to wait for the $D/N-\alpha+1 = 5 - 4 + 1 = 2$ fastest devices in each group, i.e., $10$ devices in total. 
\end{example}

Note that in the previous example the $2$ fastest devices in each group are not necessarily among the $10$ fastest devices globally. This means that the straggler mitigation capability of CodedPaddedFL with grouping may be lower than that of CodedPaddedFL with no grouping. As we will show  numerically in Section~\ref{sec:numericalResults}, the trade-off of slightly reduced straggler mitigation for much lower values of $\alpha$\textemdash thereby much lower initial communication load and lower decoding complexity at the central server\textemdash can reduce the overall latency.

\section{Coded Secure Aggregation}
\label{eq:CodedSecAgg}

In this section, we present a coding scheme, referred to as CodedSecAgg, for mitigating the effect of stragglers in FL that increases the privacy level of traditional FL schemes and CodedPaddedFL by preventing the central server from launching a model inversion attack. The higher level of privacy compared to CodedPaddedFL is achieved at the expense of a higher training time.

As with CodedPaddedFL, CodedSecAgg can be divided into two phases. First, the devices use Shamir's  SSS (see \cref{sec:shamir}) with parameters $(D,k')$ to share their local data with other devices in the network. This introduces redundancy of the data which can be leveraged for straggler mitigation. At the same time, Shamir's SSS guarantees that any subset of less than $k'$ devices does not learn anything about the local datasets of other devices. In the second phase, the devices perform gradient descent on the SSS encoded data and send their results to the central server. The central server can decode the received results from any $k'$ devices to obtain the aggregated gradient, thereby providing resiliency against up to $D-k'$ stragglers. At the same time, the central server does not gain access to any local gradient and a model inversion attack is prevented.%

\subsection{Phase 1: Data Sharing}

The devices use Shamir's  SSS with parameters $(D,k')$ to encode both $\bm G_i^{(1)}$ and ${\bm X^{(i)\top}} \bm X^{(i)}$ into $D$ shares. Let $\bigl\{\bm{\mathsf{R}}_{i,1}^\mathrm{G},\dots,\bm{\mathsf{R}}_{i,k'-1}^\mathrm{G}\bigr\}$ and $\bigl\{\bm{\mathsf{R}}_{i,1}^\mathrm{X},\dots,\bm{\mathsf{R}}_{i,k'-1}^\mathrm{X}\bigr\}$ be two sets of $k'-1$ independent and uniformly distributed matrices. Device $i$ encodes $\bm G_i^{(1)}$ together with  $\bigl\{\bm{\mathsf{R}}_{i,1}^\mathrm{G},\dots,\bm{\mathsf{R}}_{i,k'-1}^\mathrm{G}\bigr\}$ into $D$ shares $\bigl\{\bm \Psi_i^{(1)},\dots,\bm \Psi_i^{(D)}\bigr\}$ and ${\bm X^{(i)\top}} \bm X^{(i)}$ together with $\bigl\{\bm{\mathsf{R}}_{i,1}^\mathrm{X},\dots,\bm{\mathsf{R}}_{i,k'-1}^\mathrm{X}\bigr\}$ into $D$ shares $\bigl\{\bm \Phi_i^{(1)},\dots,\bm \Phi_i^{(D)}\bigr\}$ using a nonsystematic $(D,k')$ Reed-Solomon code. Subsequently, each device sends one share of each encoding to each of the other $D-1$ devices. More precisely, device $i$ sends $\bPsi_i^{(j)}$ and $\bPhi_i^{(j)}$ to device $j$.

Once the data sharing is completed, device $i$ has $\bigl\{\bm \Psi_1^{(i)},\dots,\bm \Psi_D^{(i)}\bigr\}$ and $\bigl\{\bm \Phi_1^{(i)},\dots,\bm \Phi_D^{(i)}\bigr\}$. Device $i$ then computes $\bm \Psi^{(i)} = \sum_{j=1}^D \bm \Psi_j^{(i)}$ and $\bm \Phi^{(i)} = \sum_{j=1}^D \bm \Phi_j^{(i)}$. It is easy to see that $\bigl\{\bm \Psi^{(1)},\dots,\bm \Psi^{(D)}\bigr\}$ 
and $\bigl\{\bm \Phi^{(1)},\dots,\bm \Phi^{(D)}\bigr\}$ correspond to applying Shamir's  SSS with parameters $(D,k')$ to $\bigl\{\sum_i\bm G_i^{(1)},\sum_i\bm{\mathsf{R}}_{i,1}^\mathrm{G},\dots,\sum_i\bm{\mathsf{R}}_{i,k'-1}^\mathrm{G}\bigr\}$ and  $\bigl\{\sum_i{\bm X^{(i)\top}} \bm X^{(i)},\sum_i\bm{\mathsf{R}}_{i,1}^\mathrm{X},\dots,\sum_i\bm{\mathsf{R}}_{i,k'-1}^\mathrm{X}\bigr\}$. Hence, due to the linearity of Shamir's SSS, the devices now obtained successfully a secret share of $\bm G^{(1)}$ and $\bm X^\top\bm X$, the first aggregated global gradient and the global dataset. This concludes the first phase.

\subsection{Phase 2: Securely Aggregated Gradient Descent}

The second phase is an iterative learning phase, in which the devices continue to exploit the linearity of Shamir's SSS by computing the gradient updates on their shares $\bm \Phi^{(i)}$ and $\bm \Psi^{(i)}$. They thereby obtain a share of the new gradient in each epoch. More precisely, in each epoch $e$, the devices compute
\begin{equation}
    \label{eq:secureupdate}
    \tilde{\bm G}_i^{(e)} = \bm \Psi^{(i)} + \bm \Phi^{(i)} \bm \epsilon^{(e)}\,.
\end{equation}

\begin{figure}
    \centering
    \includegraphics[width=\columnwidth]{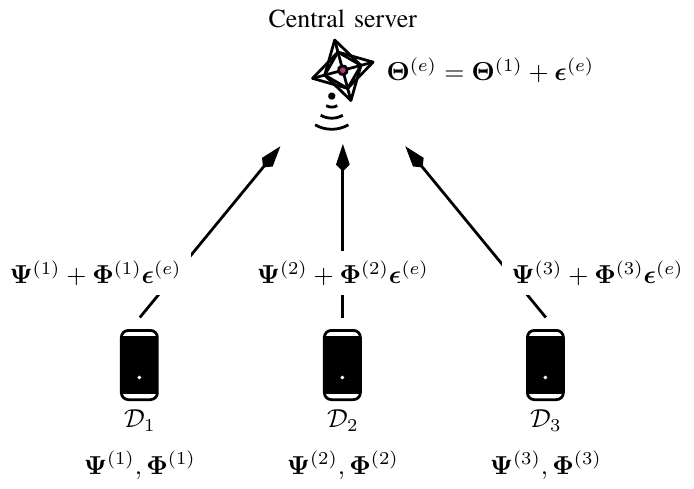}
    \caption{An example showcasing an epoch of CodedSecAgg. The system consists of $D=3$ devices and a central server. Each device has access to one share of the global dataset.}
    \label{fig:secureaggexample}
\end{figure}

In epoch $e$, the $k'$ fastest devices to finish their computation send their computed update $\tilde{\bm G}_i^{(e)}$ to the central server, which  can decode the SSS to obtain the aggregated gradient $\bm G^{(e)}$ for that epoch. At the same time, the aggregated gradient is the only information the central server\textemdash and any set of less than $k'$ colluding devices\textemdash obtains. In the first phase, the local datasets are protected by the SSS from any inference, and in the second phase, only shares of the aggregated gradients are collected, which do not leak any information that the central server was not supposed to learn\textemdash the central server is supposed to learn the aggregated gradient in each epoch and there is no additional information the central server learns. We illustrate CodedSecAgg in \cref{fig:secureaggexample}.

\begin{remark}
In order to guarantee that the central server obtains the correct model update after decoding the SSS, the devices have to modify the multiplication of fixed-point numbers: in the SSS, we interpret the fixed-point numbers as integers from $\mathbb Z_{\langle k \rangle}$. As described in \cref{sec:priv_op_fp}, multiplying two fixed-point numbers involves an integer multiplication with subsequent scaling to retain the precision of the datatype. However, it is not guaranteed that the decoding algorithm of the SSS will yield the desired result when the devices apply scaling after the integer multiplication. Whenever a wrap-around happens, i.e., the result of an integer operation exceeds $2^{k-1}$ or $-2^{k-1}$, which is expected to happen during the encoding of the SSS, the subsequent scaling distorts the arithmetic of the SSS. To circumvent this phenomenon and guarantee correct decoding of the global aggregate, we postpone the scaling and apply it after the decoding of the SSS at the central server. To this end, the range of integers we can represent has to be increased in accordance with the number of fractional bits used, i.e., the devices have to perform integer operations in $\mathbb Z_{\langle k + f\rangle}$, to guarantee that no overflows occur due to the postponed scaling. This entails an increase in computation and communication complexity of a factor $\frac{k+f}{k}$ compared to the case where no secret sharing is used. Furthermore,  \cref{eq:secureupdate} involves the addition of  the result of a multiplication and a standalone matrix $\bm\Psi^{(i)}$. In order to perform a correct scaling after the decoding, we have to artificially multiply the standalone matrix $\bm\Psi^{(i)}$ with the identity matrix. This can be done efficiently by multiplying $\bm G_i^{(1)}$ with $2^f$ prior to encoding it into $\{\bm\Psi^{(1)}_i,\ldots,\bm\Psi^{(D)}_i\}$.

Reed-Solomon codes are defined over finite fields. Thus, the operations in CodedSecAgg need to be performed over a finite field. To this end, for a fixed-point representation using $k$ bits (see Section~\ref{sec:Fixed-Point Numbers}), we consider a finite field of order $q>2^{k+f}$, where $q$ is a prime number. The mapping between the integers corresponding to the $k$-bit fixed-point representation is done as follows: we map integers $0$ to $2^{k+f-1}-1$ to the first $2^{k+f-1}$ elements of the finite field and $-1\mapsto q-1$, $-2\mapsto q-2$, and so on.
\end{remark}

\subsection{Complexity}
The complexity analysis of CodedSecAgg is almost equivalent to that of CodedPaddedFL (see~\cref{sec:paddedFLComplexity}). In particular, the complexity of CodedSecAgg's learning phase  is identical to that of  CodedPaddedFL and conventional FL. We consider now the sharing phase. In the sharing phase, device $i$ has to upload its $2\left(D-1\right)$ shares $\{\bm \Psi_i^{(j)}|j\in[D],j\neq i\}$ and $\{\bm \Phi_i^{(j)}|j\in[D],j\neq i\}$ and download the $2\left(D-1\right)$ shares $\{\bm \Psi_j^{(i)}|j\in[D],j\neq i\}$ and $\{\bm \Phi_j^{(i)}|j\in[D],j\neq i\}$. Furthermore, each device has to add $D$ matrices twice. Therefore, each device uploads and downloads $\left(D - 1\right)d\left(\frac{d+1}{2}+c\right)$ elements from $\fxp{k}{f}$ and performs $\left(D - 1\right)d\left(\frac{d+1}{2}+c\right)$ additions (MAC operations where one of the factors is set to $1$).

\subsection{Grouping for Coded Secure Aggregation}
For the proposed CodedSecAgg, the  communication cost entailed by the sharing phase and the decoding cost at the central server increase with the number of devices $D$. Similar to CodedPaddedFL, we can reduce these costs while preserving 
straggler mitigation and secure aggregation by grouping the devices into  groups and applying CodedSecAgg in each group.
However,  applying directly the above-described CodedSecAgg locally in each group would leak information about the data of subsets of devices. In particular, the central server would learn the aggregated gradients in each group instead of only the global gradient. To circumvent this problem, we introduce a hierarchical structure on the groups. Specifically, only one group, referred to as the master group,  sends updates to the central server directly. This group collects the model updates of the other groups and aggregates them before passing the global aggregate  to the central server. To prevent a communication bottleneck at the master group, we avoid all devices  sending  updates directly to this group by dividing the communication into multiple hierarchical steps. At each step, we collect intermediate aggregates at fewer and fewer groups until all group updates are aggregated at the master group.

The proposed algorithm is as follows. We group devices into $N$ disjoint groups. Contrary to CodedPaddedFL, where the groups may be of different size, here we require equally-sized groups, i.e.,  $N$ divides $D$, as the inter-group communication requires each device in a group communicating with a unique device in another group and no two devices communicating with the same device. Furthermore, we require the number of devices in each group to be at least $k'$. For notational purposes, we assign each group an identifier $j\in [N]$, and each device in a group is assigned an identifier $i \in [D/N]$ which is used to determine which shares each device receives from the other devices in its group in phase one of the algorithm. Let $\bm G_{j,i}^{(1)}$ be the first gradient of device $i$ in group $j$ and  ${\bm X^{(j,i)\top}} \bm X^{(j,i)}$  its data. Similar to the above-described scheme, device $i$ in group $j$ applies Shamir's $(D/N,k')$ SSS on $\bm G_{j,i}^{(1)}$ and ${\bm X^{(j,i)\top}} \bm X^{(j,i)}$ to obtain $D/N$ shares $\bigl\{\bm \Psi_{j,i}^{(1)},\dots,\bm \Psi_{j,i}^{(D/N)}\bigr\}$ and $D/N$ shares $\bigl\{\bm \Phi_{j,i}^{(1)},\dots,\bm \Phi_{j,i}^{(D/N)}\bigr\}$, respectively. Device $i$ sends shares $\bm \Psi_{j,i}^{(i')}$ and $\bm \Phi_{j,i}^{(i')}$ to all other devices $i'\in [D/N]\backslash i$ in the group. Device $i'$ then computes $\bm \Psi_{j}^{(i')} = \sum_{i\in[D/N]} \bm \Psi_{j,i}^{(i')}$ and $\bm \Phi_{j}^{(i')} = \sum_{i\in[D/N]} \bm \Phi_{j,i}^{(i')}$.

Let $\tilde{\bm G}_{j,i}^{(e)} = \bm \Psi_{j}^{(i)} + \bm \Phi_{j}^{(i)}\bm \epsilon^{(e)} $, $j\in[N]$, $i\in[D/N]$, be the model update of device $i$ in group $j$ at epoch $e$ equivalently to \cref{eq:secureupdate} and  $\bm \bar{\bm G}^{(e)}_j$  the aggregated gradient of group $j$ in epoch $e$. Similar to the above-described CodedSecAgg scheme, $\bigl\{\tilde{\bm G}_{j,1}^{(e)},\dots,\tilde{\bm G}_{j,D/N}^{(e)}\bigr\}$ is the result of applying Shamir's $(D/N,k')$ SSS on $\bm \bar{\bm G}^{(e)}_j$. In order to prevent the central server from learning $\bm \bar{\bm G}^{(e)}_j$ for any $j$, the key idea is to compute $D/N$ shares of $\bm G^{(e)} = \sum_{j\in[N]}\bm \bar{\bm G}^{(e)}_j$ in the master group. Devices in the master group can then send their shares of $\bm G^{(e)}$ to the central server, which can decode the SSS from any $k'$ shares to obtain  $\bm G^{(e)}$. Note that we want to prevent the central server inferring any of the individual $\bm \bar{\bm G}^{(e)}_j$. We can achieve this by letting device $i$ in the master group compute $\sum_j \tilde{\bm G}_{j,i}^{(e)}$. As each $\tilde{\bm G}_{j,i}^{(e)}$ is one out of $D/N$ shares of $\bar{\bm G}_{j}^{(e)}$, $\bigl\{\sum_j \tilde{\bm G}_{j,1}^{(e)},\dots,\sum_j \tilde{\bm G}_{j,D/N}^{(e)}\bigr\}$ is the result of applying Shamir's SSS on $\sum_{j\in[N]} \bar{\bm G}_{j}^{(e)} = \bm G^{(e)}$. %

The model updates $\tilde{\bm G}_{j,i}^{(e)}$ are not sent directly to device $i$ in the master group, to avoid a communication bottleneck in the master group. In particular, we divide the communication round into multiple steps.  Assume for simplicity, and without loss of generality, that the master group is group one. We proceed as follows. %
In the first step, each device $i$ in group $j \in \{j' \in [N]~|~j'\bmod2 = 0\}$ sends $\tilde{\bm G}_{j,i}^{(e)}$ to device $i$ in group $j - 1$, which adds its own share and the received one, i.e., the devices in group $j-1$ obtain a share of $\bm \bar{\bm G}^{(e)}_{j-1} + \bm \bar{\bm G}^{(e)}_j$.
In the second step, each device $i$ in group $j \in \{j' \in [N]~|~j'\bmod4 = 3\}$ sends $\bm \tilde{\bm G}^{(e)}_{j,i} + \bm \tilde{\bm G}^{(e)}_{j+1,i}$ to device $i$ in group $j-2$ which again aggregates the received  shares and its own. Devices in group $j-2$ now have obtained a share of $\bm \bar{\bm G}^{(e)}_{j-2} + \bm \bar{\bm G}^{(e)}_{j-1}$ + $\bm \bar{\bm G}^{(e)}_{j} + \bm \bar{\bm G}^{(e)}_{j+1}$. Generally, in step $s$, each device $i$ in group
\begin{equation}
    \label{eq:grouptransmitter}
    j\in \{j' \in [N]~|~j'\bmod 2^s = 2^{s-1} + 1\bmod 2^s\}\end{equation}
sends
\begin{equation}
    \label{eq:groupingaggregate}
    \sum_{j'=j}^{\min\{j+2^{s-1}-1, N\}} \bm \tilde{\bm G}^{(e)}_{j',i}
\end{equation}
to device $i$ in group $j-2^{s-1}$. We continue this process until the devices in group one (the master group) have obtained shares of the global gradient $\bm G^{(e)} = \sum_j\bm \bar{\bm G}^{(e)}_j$. In total, $\lceil\log_2(N)\rceil$ steps are needed to reach this goal.

We illustrate the inter-group communication with the following example.
\begin{figure}
  \centering
  \includegraphics[width=\columnwidth]{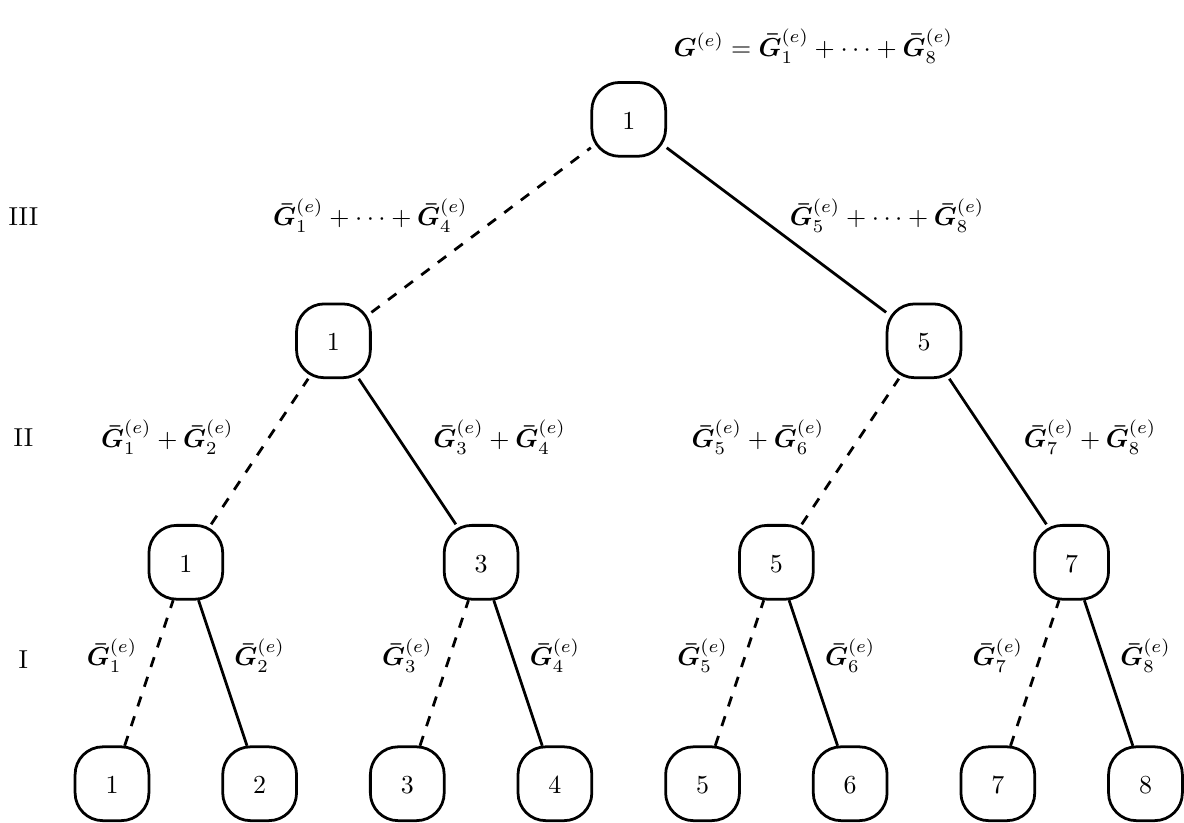}
  \caption{An example of the inter-group communication for a network with $N=8$ groups. Different layers correspond to the $\lceil\log_2(N)\rceil = 3$ communication steps, while the label of each node is the group identifier. A solid line between node $i$ and node $j$ represents a physical transmission from devices in group $i$ to devices in group $j$, whereas dashed lines represent data already available at the end node.}
  \label{fig:example}
\end{figure}
\begin{example}
  Consider a network with $N=8$  groups as depicted in \cref{fig:example}, where the groups are numbered $1$ to $8$ and represented by squares. In the first step,  devices in group $2$ send
  their shares of $\bm  \bar{\bm G}^{(e)}_2$ to devices in group  $1$, devices in group $4$ their shares of $\bm  \bar{\bm G}^{(e)}_4$ to group  $3$, devices in group $6$ their shares of $\bm  \bar{\bm G}^{(e)}_6$ to group $5$,
  and devices in group $8$ their shares of $\bm  \bar{\bm G}^{(e)}_8$ to group $7$,  which is illustrated by the solid  lines. After the first step, each device  in group $1$ has
  access to a share of $\bm \bar{\bm G}^{(e)}_1 +  \bm \bar{\bm G}^{(e)}_2$, devices in group $3$ have shares of 
  $\bm \bar{\bm G}^{(e)}_3  + \bm \bar{\bm G}^{(e)}_4$,  and so forth. In  the second step, devices from  group $3$ send
  their shares of $\bm \bar {\bm G}^{(e)}_3  + \bm \bar{\bm G}^{(e)}_4$ to devices in  group $1$ and devices  from group $7$ their  shares of $\bm \bar {\bm G}^{(e)}_7  + \bm \bar{\bm G}^{(e)}_8$ to group
  $5$. In the last step, the devices in group $5$ send their shares of $\bm \bar {\bm G}^{(e)}_5  + \bm \bar{\bm G}^{(e)}_6 + \bm \bar {\bm G}^{(e)}_7 + \bm \bar {\bm G}^{(e)}_8 $ to the devices in group $1$ which now have a share of the global aggregate $\bm G^{(e)}$.
    
  Notice that  there is at most one  solid incoming and outgoing edge  at each node. This means  that at any
  step devices receive at most one message and send at most one message to devices in another group. This way we
  avoid a congestion of the network and can collect the shares of the aggregates efficiently in group 1. Furthermore, although we picked $N$ to be a power of $2$, \cref{eq:grouptransmitter} and \cref{eq:groupingaggregate} hold for any $N$ that divides $D$.
\end{example}

At any step, each device has access to at most one share of any $\bm \bar{\bm G}^{(e)}_j$. Note that shares from different groups encode different gradients and can not be used together to extract any information. As a result, the privacy is not impaired by the grouping as we still need $k'$ colluding devices to decode any SSS, while the central server is only able to decode the global aggregate.

Both the grouping and the inter-group communication are detrimental to straggler mitigation compared to CodedSecAgg with only one group. %
However, as we show in the next section, the reduced decoding cost at the central server and the reduced communication cost in the first phase compensate for the reduced straggler mitigation.

\section{Numerical Results}
\label{sec:numericalResults}

We simulate an FL network in which devices want to collaboratively train on the MNIST \cite{Cun10} and Fashion-MNIST \cite{Xia17} datasets, i.e., we consider the application of the proposed schemes to a classification problem. To do so, we preprocess the datasets using kernel embedding via Python's radial basis function sampler of the sklearn package (with $5$ as kernel parameter and $2000$ features). %
We divide the datasets into training and test sets. Furthermore, the training set is sorted according to the labels to simulate non-identically distributed data before it is divided into $D$ equally-sized batches which are assigned without repetition to the $D$ devices. We use $k=48$ bits to represent fixed-point numbers with a resolution of $f=24$ bits in CodedPaddedFL and CodedSecAgg, whereas we use $32$-bit floating point numbers to represent the data for the schemes we compare with, i.e., conventional FL, the scheme in \cite{Pra21}, and LightSecAgg. For our proposed schemes, we assume that the computation of the first local gradient and $\bm X^{(i)\top}\bm X^{(i)}$ happens offline because no interaction is required by the devices to compute those. %
We sample the setup times $\mathsf{\Lambda}_i$ at each epoch and assume that they have an expected value of $50$\% of the deterministic computation time. In particular, device $i$ performing $\rho_i$ MAC operations at each epoch yields $\eta_i = \frac{2\tau_i}{\rho_i}$.
For the communication between the central server and the devices, we assume they use the LTE Cat 1 standard for IoT applications, which means that the corresponding rates are $\gamma^{\mathsf{d}}=10$ Mbit/s and $\gamma^{\mathsf{u}}=5$ Mbit/s. The probability of transmission failure is $p_i=0.1$ and we add a $10$\% header overhead to all transmissions. %
For the learning, we use a regularization parameter $\lambda=9\times10^{-6}$ and an initial learning rate of $\mu = 6.0$, which is updated as $\mu\leftarrow0.8\mu$ at epochs $200$ and $350$.

\subsection{Coded Federated Learning}
\begin{figure}
    \centering
    \begin{subfigure}[b]{0.48\textwidth}
        \centering
        \includegraphics[width=1\columnwidth]{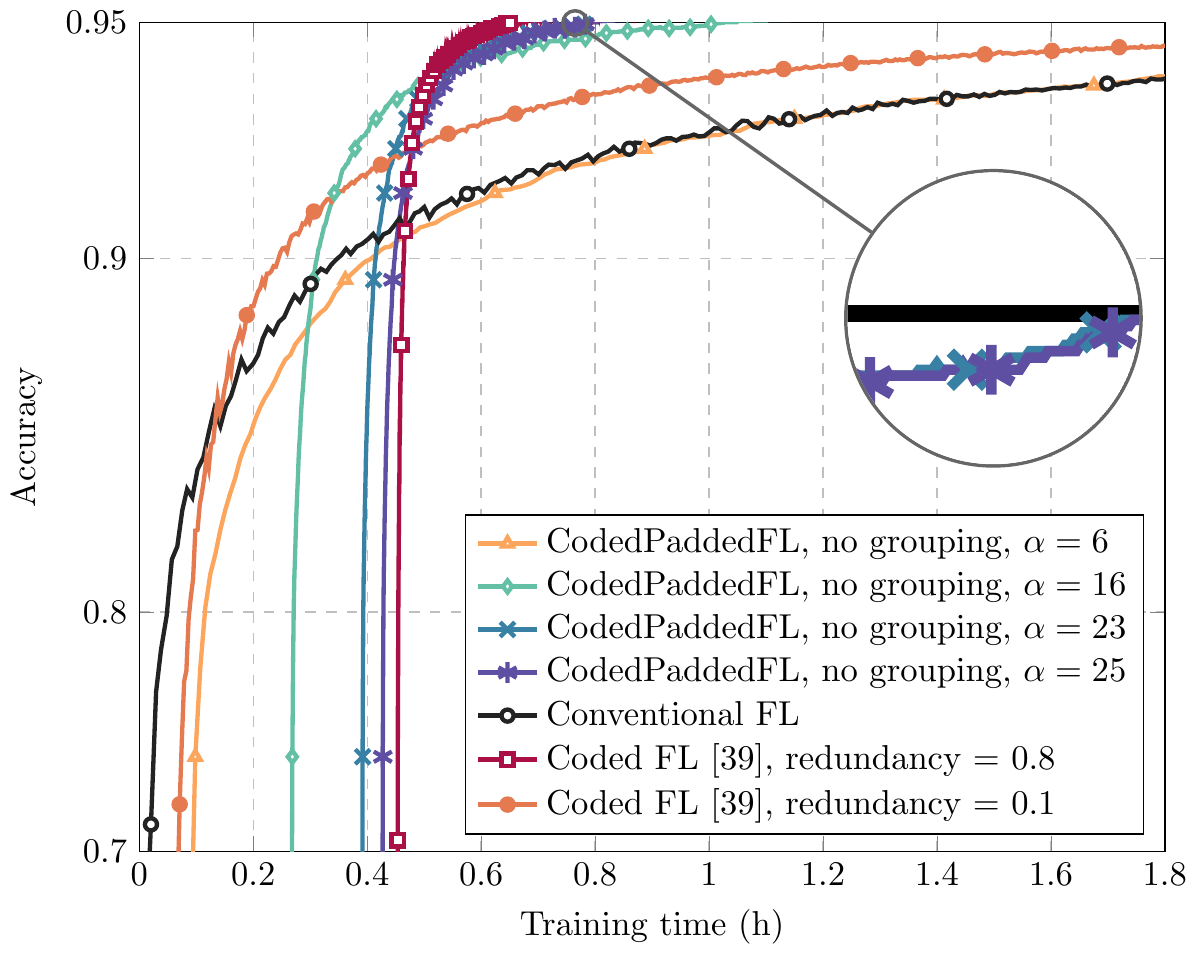}
        \caption{MNIST dataset}
        \label{Fig: TrainingTimeVSaccuracy_nu0.1}
    \end{subfigure}
    \hfill
    \vspace{0.25cm}
    \begin{subfigure}[b]{0.48\textwidth}
        \centering
        \includegraphics[width=1\columnwidth]{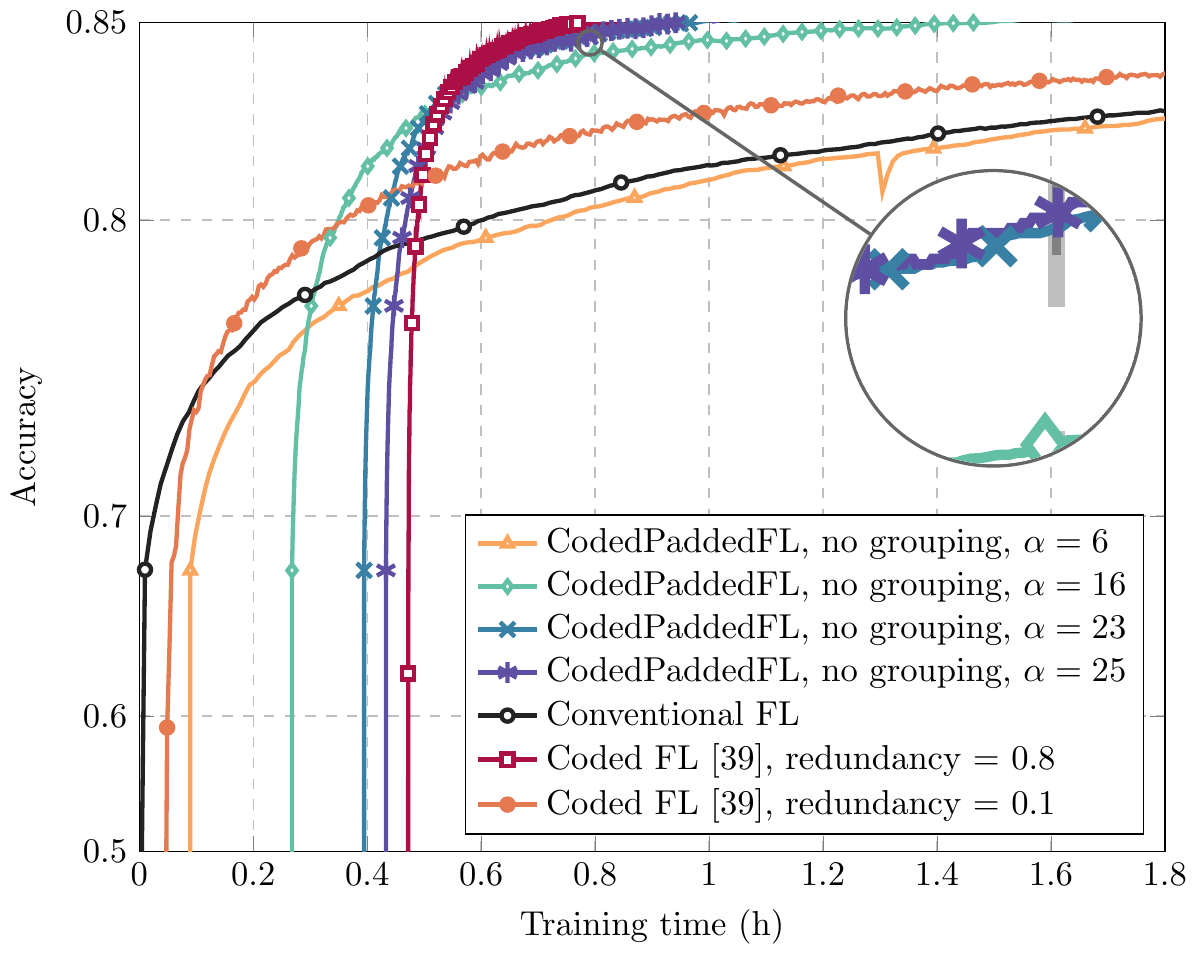}
        \caption{Fashion-MNIST dataset}
        \label{Fig: FMNIST_TrainingTimeVSaccuracy_nu0.1}
    \end{subfigure}
    \caption{Training time for the proposed CodedPaddedFL with different values of $\alpha$, the coded FL scheme in \cite{Pra21}, and conventional FL.}
\end{figure}

We first consider a network with $D=25$ devices. We model the heterogeneity by varying the MAC rates $\tau_i$ across devices. In particular, we have $10$ devices with a MAC rate of $25 \cdot 10^6$ MAC/s, $5$ devices with $5 \cdot 10^6$, $5$ with $2.5 \cdot 10^6$, and the last $5$ with $1.25 \cdot 10^6$, whereas the central server has a MAC rate of $8.24 \cdot 10^{12}$ MAC/s. These MAC rates are chosen in accordance with the performance that can be expected from devices with chips from the Texas Instruments TI MSP430 family\cite{TIchip}. For the conventional FL training, we perform mini-batch gradient descent where we use a fifth of the data at each epoch. We chose the mini-batch size as a compromise between the extreme cases: a low mini-batch size does not allow for much parallelization whereas a large mini-batch size might be exceeding the parallelization capabilities of the devices and thereby slow the training down. Note that for CodedPaddedFL, we train on $\bm X^{(i)\top}\bm X^{(i)}$ for which it does not give any benefits to train on mini-batches.

In \cref{Fig: TrainingTimeVSaccuracy_nu0.1}, we plot the accuracy over the training time on the MNIST dataset for the proposed CodedPaddedFL with no grouping, i.e., for $N=1$, for different values of $\alpha\in\{6, 16, 23, 25\}$, conventional FL, and the scheme in \cite{Pra21}. Note that $\alpha = 25$ corresponds to a replication scheme where all devices share their padded data with all other devices. By the initial offsets in the plot, we can see that the encoding and sharing, i.e., phase one, takes longer with increasing values of $\alpha$. However, the higher straggler mitigation capabilities of high values of $\alpha$ result in steep curves. Our numerical results show that the optimal value of $\alpha$ depends on the target accuracy. For the considered scenario, $\alpha = 23$ reaches an accuracy of $95$\% the fastest. Conventional FL has no initial sharing phase, so the training can start right away. However, the lack of straggler mitigation capabilities result in a slow increase of accuracy over time. %
For an accuracy of $95$\%, CodedPaddedFL yields a speed-up factor of $6.6$ compared to conventional FL. For levels of accuracy below $90$\%, conventional FL performs best and there are also some $\alpha$, such as $\alpha = 6$, where the performance of CodedPaddedFL is never better than for conventional FL.

The scheme in \cite{Pra21} achieves speed-ups in training time by trading off the users' data privacy. In short, in this scheme devices offload computations to the central server through the parity data to reduce their own epoch times. The more a device is expected to straggle, the more it offloads to the central server. To quantify how much data is offloaded, the authors introduce a parameter called redundancy which lies between $0$ and $1$. A low value of redundancy means little data is offloaded and thereby leaked, whereas a high value of redundancy means that the central server does almost all of the computations and results in a high information leakage. We consider two levels of redundancy for our comparison, $0.1$ and $0.8$. We can see that for a redundancy of $0.1$ our scheme outperforms the scheme in \cite{Pra21} for significant levels of accuracy whereas CodedPaddedFL is only slightly slower for a redundancy of $0.8$. Note, however, that a redundancy of $0.8$ means that the devices offload almost all of the data to the central server, which not only leaks the data but also transforms the FL problem into a centralized learning problem.

In \cref{Fig: FMNIST_TrainingTimeVSaccuracy_nu0.1}, we plot the accuracy over training time for the Fashion-MNIST dataset with no grouping, i.e., for $N=1$. We observe a similar qualitative performance. In this case,  $\alpha=25$ is the value for which CodedPaddedFL achieves fastest an accuracy of $85$\%, yielding a speed-up factor of $9.2$ compared to conventional FL. For a target accuracy between $80$\% and $85$\%, CodedPaddedFL with different $\alpha < 25$ performs  best.

\subsection{Client Drift}

\begin{figure}
    \centering
    \begin{subfigure}[b]{0.48\textwidth}
        \centering
        \includegraphics[width=1\columnwidth]{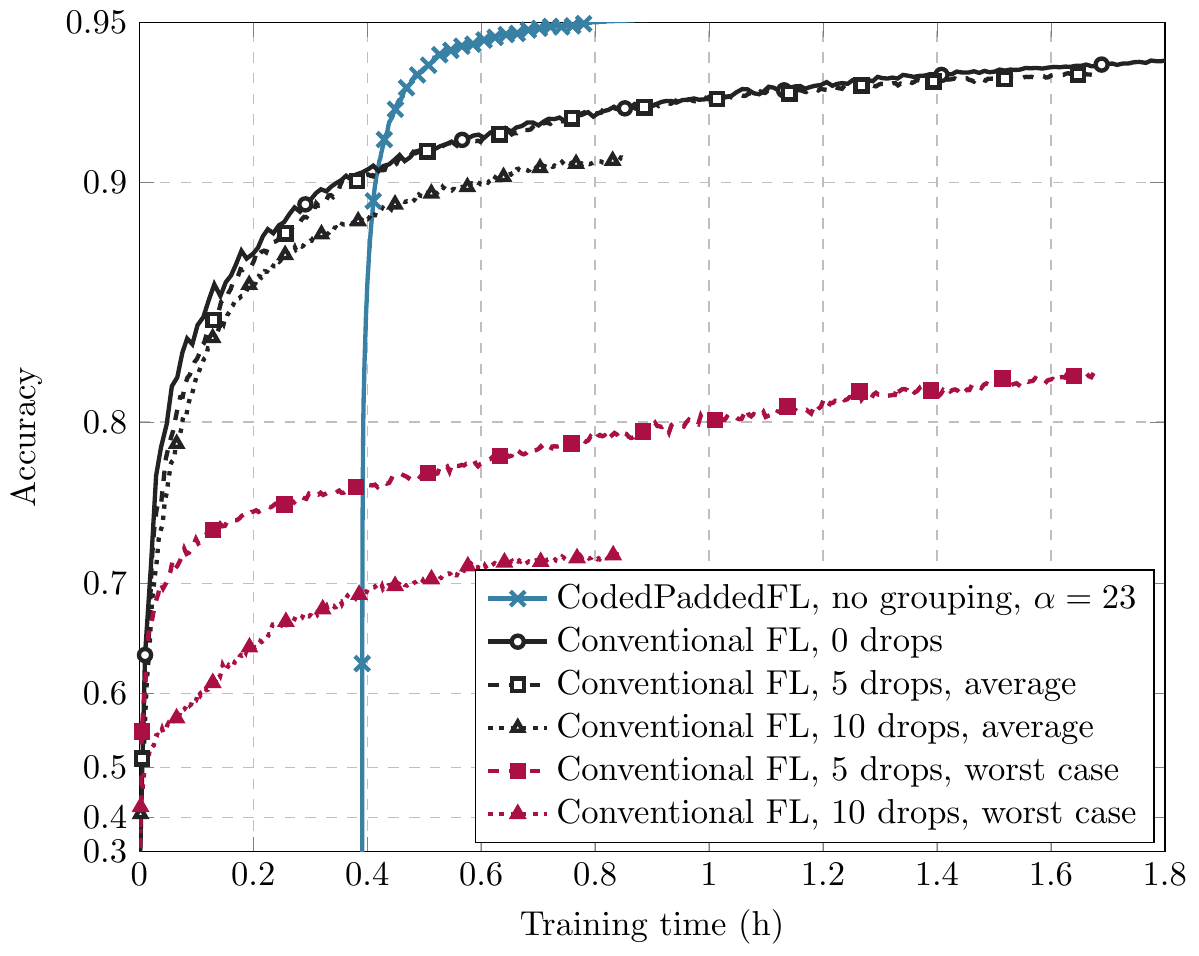}
        \caption{MNIST dataset}
        \label{Fig: TrainingTimeVSaccuracy_nu0.15}
    \end{subfigure}
    \hfill
    \vspace{0.25cm}
    \begin{subfigure}[b]{0.48\textwidth}
        \centering
        \includegraphics[width=1\columnwidth]{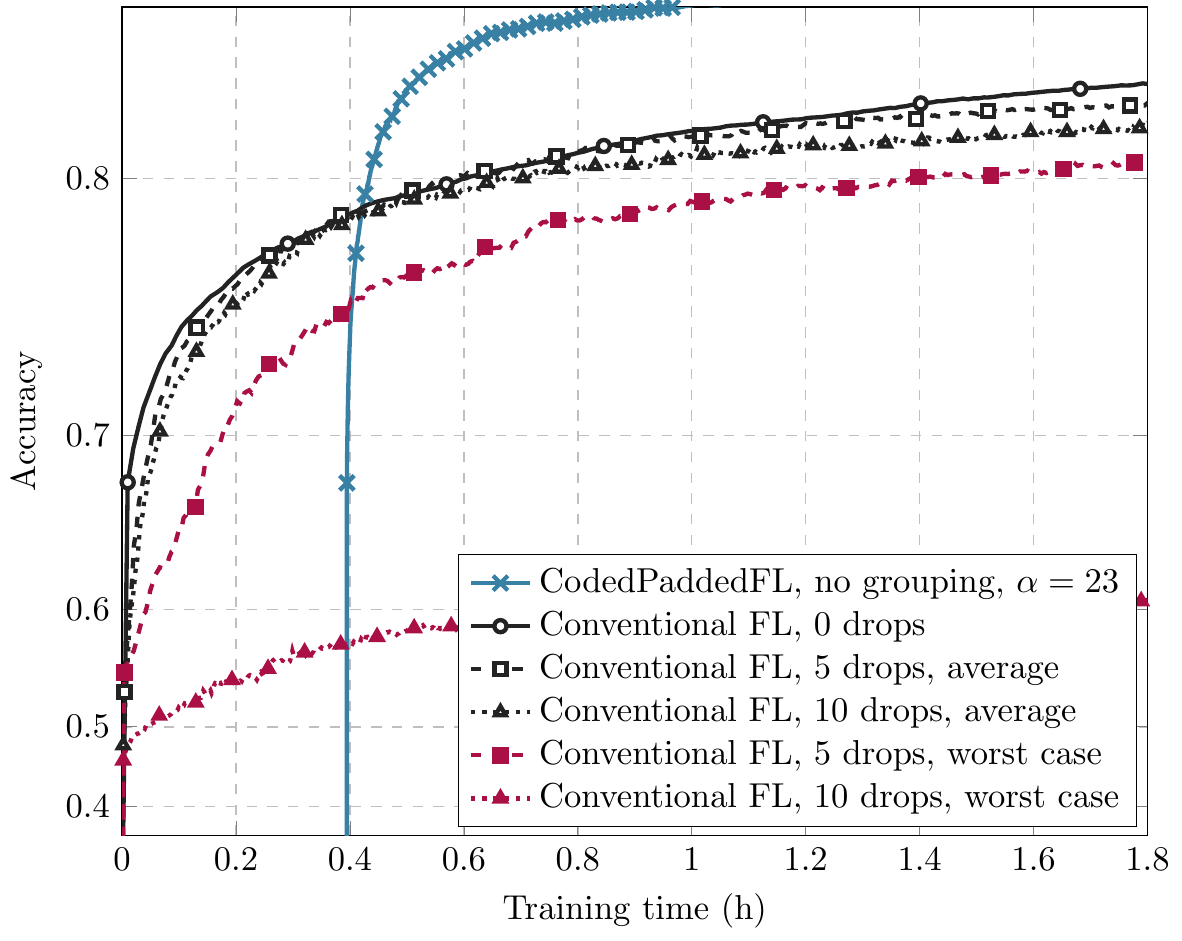}
        \caption{Fashion-MNIST dataset}
        \label{Fig: FMNIST_TrainingTimeVSaccuracy_nu0.15}
    \end{subfigure}
    \caption{Training time for the proposed CodedPaddedFL with $\alpha = 23$ and  conventional FL with a subset of the fastest devices.}
\end{figure}

In \cref{Fig: TrainingTimeVSaccuracy_nu0.15,Fig: FMNIST_TrainingTimeVSaccuracy_nu0.15}, we compare the performance of CodedPaddedFL with $N=1$ (i.e., no grouping) for $\alpha = 23$ with that of conventional FL where the $5$ or $10$ slowest devices are dropped at each epoch for the MNIST and Fashion-MNIST datasets, respectively. For  conventional FL, we plot the average performance and the worst-case performance. Dropping devices in a heterogeneous network with strongly non-identically distributed data can have a big impact on the accuracy. This is highlighted in the figures; while dropping devices causes a limited loss in accuracy on average, in some cases the loss  is significant. For the MNIST dataset, in the worst simulated case, the accuracy reduces to $82.4$\% for $5$ dropped devices  and to $72.1$\% for $10$ dropped devices (see \cref{Fig: TrainingTimeVSaccuracy_nu0.15}), underscoring the client drift phenomenon. For the Fashion-MNIST dataset, the accuracy reduces to  $82.7$\% and $61.5$\% for $5$ and $10$ dropped devices, respectively (see \cref{Fig: FMNIST_TrainingTimeVSaccuracy_nu0.15}).
The proposed CodedPaddedFL outperforms conventional FL in all cases; CodedPaddedFL has the benefit of dropping the slow devices in each epoch while not suffering from a loss in accuracy due to the redundancy of the data across the devices (see Proposition~\ref{prop:Prop}).

\subsection{Grouping}
\begin{figure}
    \centering
    \includegraphics[width=1\columnwidth]{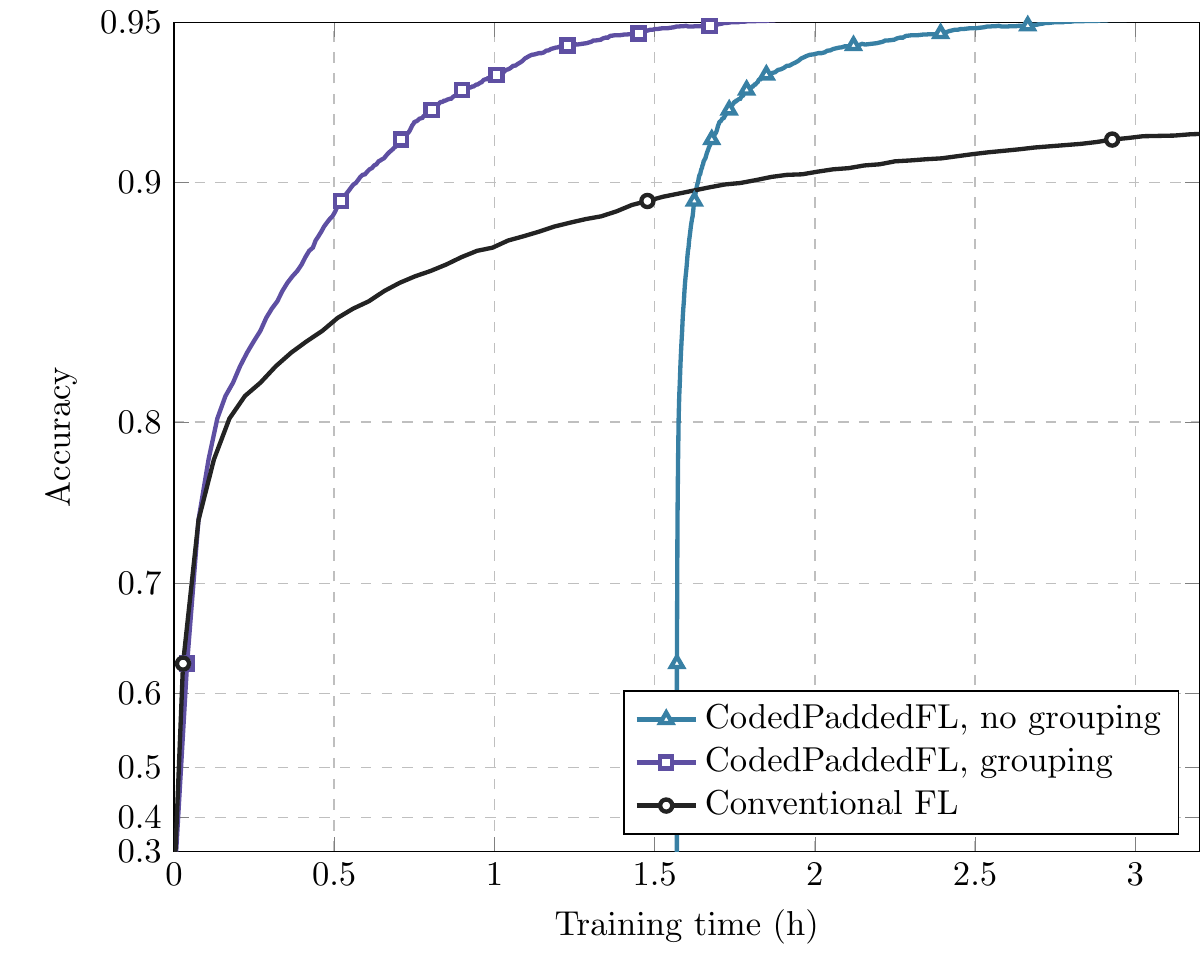}
    \caption{Training on the MNIST dataset with CodedPaddedFL, with and without grouping the $D=120$ devices.\label{fig:grouping}}
\end{figure}
The advantages of grouping the devices when training on the MNIST dataset are demonstrated in \cref{fig:grouping}. We now consider a network with $D=120$ devices and draw their MAC rates uniformly at random from the set of available rates ($25$, $5$, $2.5$, and $1.25 \cdot 10^6$ MAC/s). For a given target accuracy, we minimize the training time over all values of $\alpha$ and number of groups $N$ with the constraint that $N$ divides $D$ to limit the search space. For the baseline without grouping, we fix $N=1$. As we can see,  grouping significantly reduces the initial communication load of CodedPaddedFL. This is traded off with a shallower slope due to slightly longer average epoch times because of the reduced straggler mitigation capabilities when grouping devices. Nevertheless, the gains from the reduced time spent in phase one are too significant, and grouping the devices reduces the overall latency significantly for $D=120$ devices.
As a result, CodedPaddedFL with grouping achieves a speed-up factor of $18$ compared to conventional FL for a target accuracy of $95$\% in a network with $D=120$ devices.

\subsection{Coded Secure Aggregation}
\begin{figure}
    \centering
    \includegraphics[trim=1.4cm 0 0 0,width=1\columnwidth]{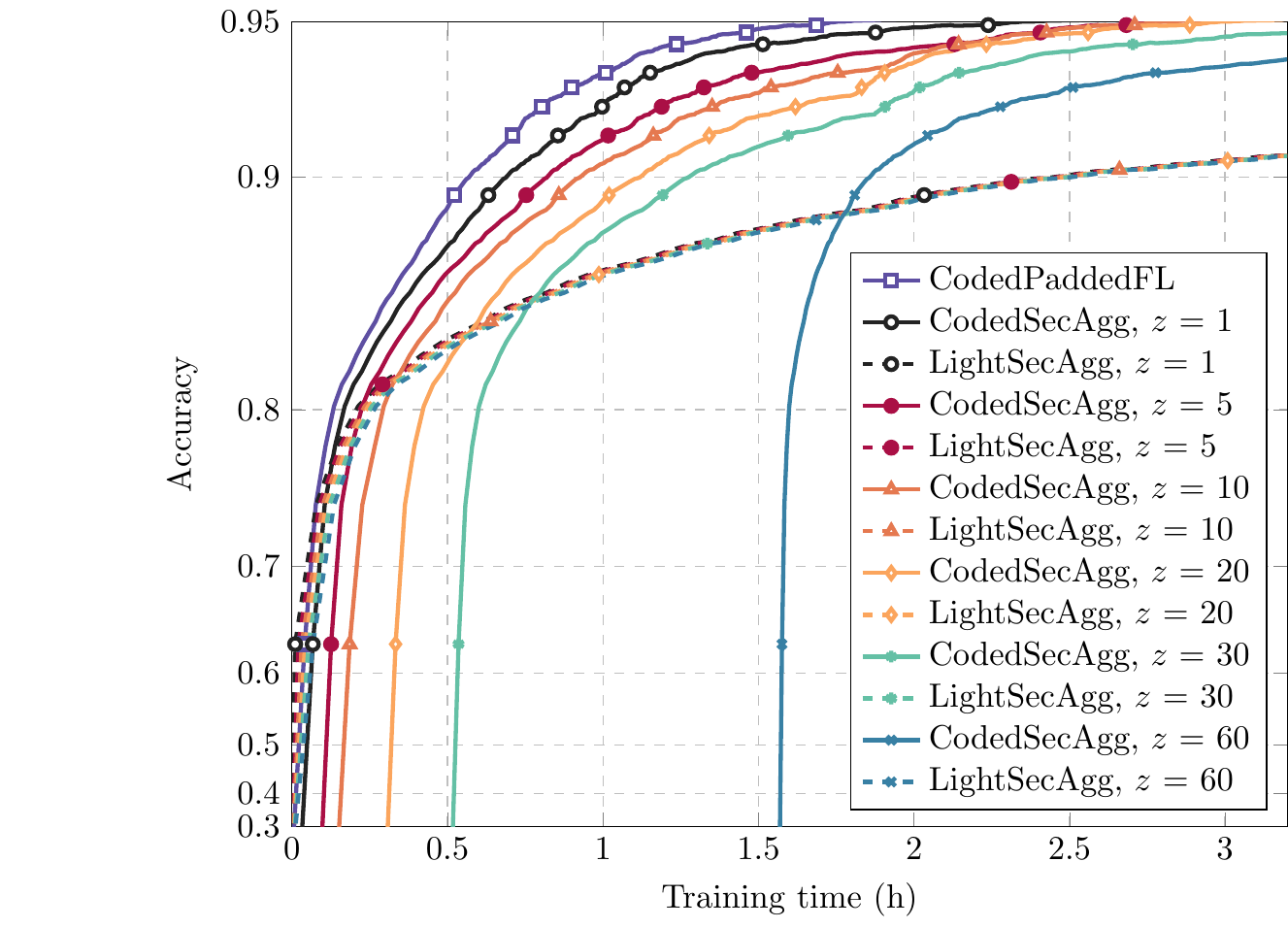}
    \caption{Training on the MNIST dataset with CodedPaddedFL and CodedSecAgg with grouping in comparison to LightSecAgg for $D=120$ devices.\label{fig:codedSecAgg}}
\end{figure}

In \cref{fig:codedSecAgg}, we plot the training over time for our proposed CodedSecAgg  and compare it with LightSecAgg for different number of colluding agents $z$ when training on the MNIST dataset. We consider a network with $D=120$ devices. As with CodedPaddedFL,  our proposed scheme again distinguishes itself by the initial offset due to the sharing of data in phase one which quickly is made up for by a much reduced average epoch time due to the straggler mitigation in phase two. As a result, our proposed CodedSecAgg achieves a speed-up factor of $18.7$ compared to LightSecAgg for a target accuracy of $95$\% when providing security against a single {malicious agent.

We notice that  CodedSecAgg  is more sensitive to an increase in the security level $z$ whereas LightSecAgg is almost unaffected regardless whether one desires to be secure against {a single malicious agent or $60$ colluding agents. A reason why our proposed scheme is more affected lies in the grouping: we require $k'>z$, i.e., we require  more than $z$ devices in each group. As a result, a high value of $z$ restricts the flexibility in grouping the devices. Nevertheless, even when we allow $60$ agents to collude, i.e., half of the devices, our scheme achieves a speed-up factor of $6.6$. Note that for $z=60$, CodedSecAgg requires at least $61$ devices per group. Given that there are only $120$ devices in total and we require each group to have the same size, there is only one group of devices, i.e., no grouping, for $z=60$.

We can also quantify the additional cost in terms of latency that secure aggregation imposes compared to CodedPaddedFL where the central server may learn the local models. For an accuracy of $95$\% on the MNIST dataset, to prevent a model inversion attack CodedSecAgg incurs a moderate additional $34$\% of latency compared to  CodedPaddedFL. %

\begin{figure}
    \centering
    \includegraphics[trim=1.4cm 0 0 0,width=1\columnwidth]{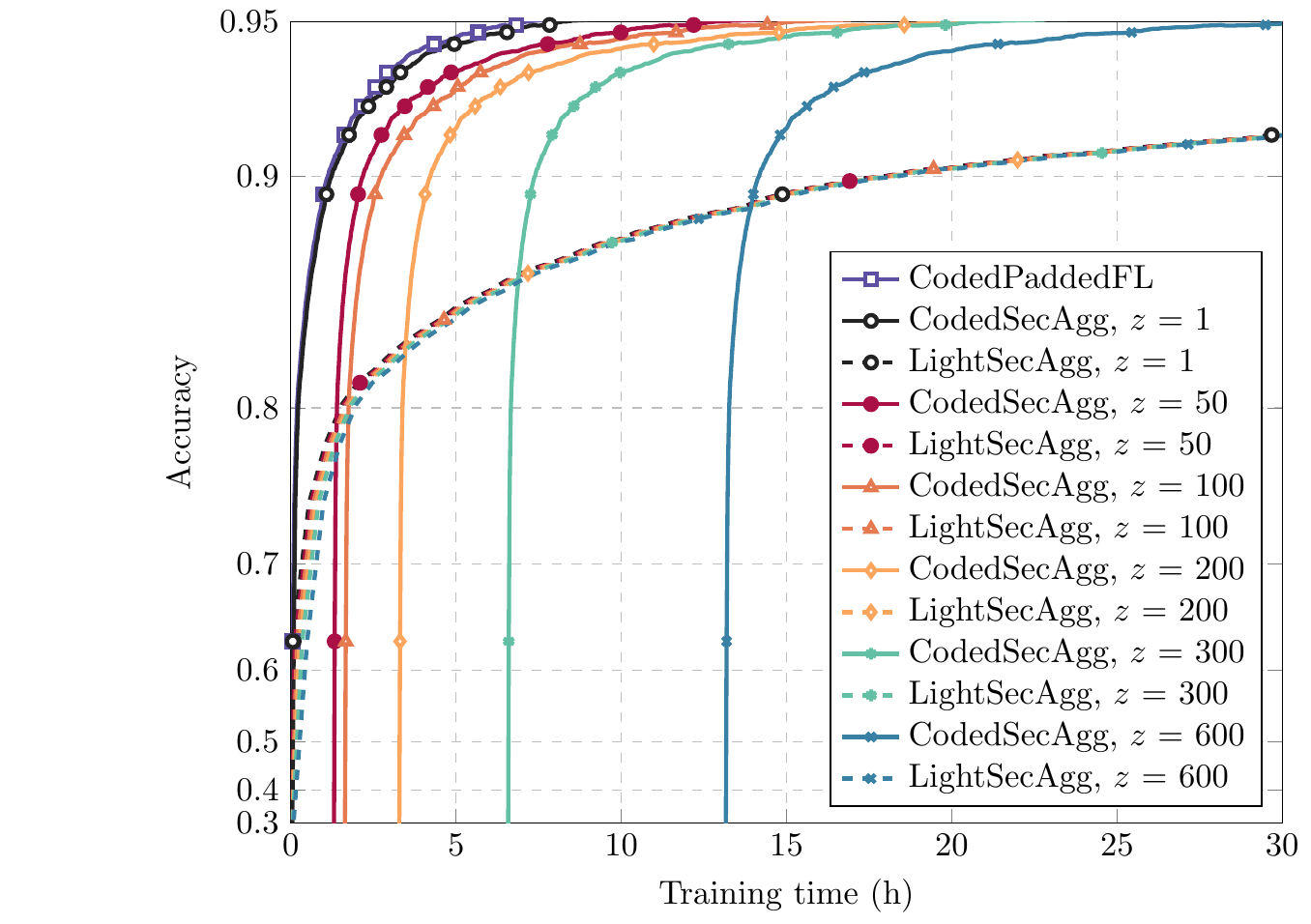}
    \caption{Training on the MNIST dataset with CodedPaddedFL and CodedSecAgg with grouping in comparison to LightSecAgg for $D=1000$ devices.\label{fig:codedSecAgg1000}}
\end{figure}

In \cref{fig:codedSecAgg1000}, we increase the number of devices in the network to $D=1000$. Qualitatively little changes compared to the scenario in \cref{fig:codedSecAgg}, which highlights the great scalability with the number of devices of our proposed scheme. However, we see a decrease in the additional relative latency due to secure aggregation. CodedSecAgg with a privacy level of $z=1$ now needs only $15$\% more latency compared to CodedPaddedFL. The higher cost of the sharing phase of CodedSecAgg compared to CodedPaddedFL becomes negligible in the long run. Due to the large number of devices, the straggler effect becomes more severe and the epoch times become longer. In comparison, the slightly longer sharing phase of CodedSecAgg is barely noticeable. Compared to LightSecAgg, CodedSecAgg achieves a speed-up factor of $10.4$ for $600$ colluding agents, whereas the speed-up factor increases to $38.9$ for a single malicious agent in the network.

We observe similar performance of CodedSecAgg on the Fashion-MNIST dataset, both compared to LightSecAgg and CodedPaddedFL.

\section{Conclusion}
We proposed two new federated learning schemes schemes, referred to as CodedPaddedFL and CodedSecAgg,
that mitigate the effect of stragglers. The proposed schemes borrow concepts from coded distributed computing to introduce redundancy across the network, which is leveraged during the iterative learning phase to provide straggler resiliency\textemdash  the central server can update the global model based on the responses of a subset of the devices. 

CodedPaddedFL and CodedSecAgg yield significant speed-up factors compared to conventional federated learning  and the state-of-the-art secure aggregation scheme LightSecAgg, respectively. Further, they converge to the global optimum and do not suffer from the client drift problem. %
While the proposed schemes are tailored to linear regression, they can be applied to nonlinear problems such as classification through kernel embedding.

\balance

\bibliographystyle{IEEEtran}
\bibliography{CFGD.bib}

\end{document}